\documentclass[10pt,journal,letterpaper,compsoc,twoside]{IEEEtran}
\usepackage{graphicx}
\usepackage{setspace}
\usepackage{times}
\usepackage{epsfig}
\usepackage{algo}
\usepackage{amsmath}
\usepackage{amssymb}
\usepackage{multirow}
\usepackage{url}
\usepackage{wrapfig}

\input{alldefin}

\begin{document}

\title{Robust Estimation of Multiple Inlier Structures}

\author{Xiang~Yang~ and~  Peter~Meer   
\IEEEcompsocitemizethanks{\IEEEcompsocthanksitem
X. Yang (Dept. of Mechanical and Aerospace  Engineering),~
P. Meer (Dept. of Electrical and Computer Engineering), 
Rutgers University, NJ 08854, USA. \protect
E-mail: \{xiang.yang@, meer@soe\}.rutgers.edu}
\thanks{}}

\markboth{Yang and Meer: Robust Estimator of Multiple Inlier Structures}
{Yang and Meer: Robust Estimator of Multiple Inlier Structures}

\IEEEcompsoctitleabstractindextext{
\begin{abstract}
The robust estimator presented in this paper processes each structure  independently.
The scales of the structures are estimated adaptively and no threshold is involved in spite of different objective functions.
The user has to specify only the number of elemental subsets for random sampling.
After classifying all the input data, 
the segmented structures are sorted by their strengths and the strongest inlier structures come out at the top. 
Like any robust estimators, this algorithm also has limitations which are described in detail.
Several synthetic and real examples are presented to illustrate every aspect of the algorithm.
\end{abstract}

\begin{IEEEkeywords}
scale estimate, structure segmentation, strength based classification, 
heteroscedasticity in computer vision.
\end{IEEEkeywords}}

\maketitle
\IEEEdisplaynotcompsoctitleabstractindextext
\IEEEpeerreviewmaketitle

\section{Introduction}
\label{sec:introduction}

\IEEEPARstart{I}n computer vision, the 
objective functions of inlier structures can be
either linear, like the estimation of planes, or nonlinear, 
such as  finding the homography between two images. 
The function can be satisfied by multiple inlier structures, and 
each structure can be corrupted by noise independently, resulting in 
different scales. 
Since all the other points are considered as outliers for a single inlier
structure, when multiple structures exist, the input data 
have lower inlier ratios.
A robust estimator detects the inlier structures from the input data,
while removing the structureless outliers.

Elemental subsets are the building blocks of the robust regression.
Each randomly chosen subset has the minimum number of points required
to estimate the parameters in the objective function.
The most used algorithm for robust fitting in the past 35 years is
the RANdom SAmple Consensus (RANSAC) \cite{fischler81}. 
Similar types of methods also exist, like PROSAC, MLESAC, Lo-RANSAC, etc.
These algorithms use different ways to generate the random sampling
and/or probabilistic relations for the
elimination of the outliers. In  paper \cite{raguram08}, a 
review of these methods was given.

In \cite{raguram13}, a universal framework for RANSAC, the Universal RANSAC (USAC),  
was introduced. However, the USAC failed in homography estimation  when a wide angle existed between two images \cite{litman15}. 
In the latter paper,
a very dense sampling of the grid was used and combined with probabilistic reasoning to obtain the inlier rate estimate. 
Only one inlier structure can be recovered.
In \cite{lavva08} RANSAC was combined with the
projection based M-estimator \cite{chen03}, 
to detect primitives such as planes, spheres, 
cylinders, etc. The 3D range data containing thousands of points
were acquired by 3D scanner and the method was tested only
for low noise cases without randomly distributed outliers.

A major drawback of RANSAC is that
the user has to give a single suitable value as the inlier scale.
Providing  too small of a scale could filter out many inlier points, 
while too large of a value bring in outliers.
In real images, for homography and fundamental matrix, the
inlier noise is usually less than 2-3 pixels,  RANSAC is successful
and sometimes the threshold is not even mentioned.
However, when the input images are scaled and/or cropped before estimation, 
the RANSAC scale may no longer be valid for a correct result. 
Problems also appear if multiple inlier structures with
different noise levels exist, or if the inlier noise varies in a sequence of images. 

In \cite{wang12}, several techniques were described 
which did not require a scale value from the user.
The proposed method derived the scale based on the $k$-th ordered absolute residual, where $k$ was 10\% of the size of the input data. Similar as in \cite{tennokoon16},
$p+2$ points were randomly selected, where $p$ was the size of the elemental subset. The $k$-th order statistics was iteratively improved to get the final estimate.
The significance of taking two additional points in a sample 
was not justified and the parameter $k$ varied largely in the experiments.
The number of inlier structures was given before the estimation, 
in order to get comparable results with the sparse subspace clustering method \cite{elhamifar13}.

Energy-based minimization approach can also be used in robust estimation, as it optimizes the quality of the entire solution in  
the Propose Expand and Re-estimate Labels (PEARL) algorithm \cite{boykov12}.
The method started with RANSAC, then followed with  alternative steps of 
expansion (inlier classification) and re-estimation to minimize the energy of the errors. 
PEARL converged to a local optimum, generally with a small number of inlier structures. A synthetic example in Figure 11 of \cite{boykov12} showed that PEARL can handle the estimation of multiple 2D lines with different Gaussian noises,
but it was not tried for real images.
The amount of outliers was relatively small in all the experiments.

The Random Cluster Model SAmpler (RCMSA) in \cite{pham14} was
similar to \cite{boykov12}, but simulated annealing was used to minimize
the overall energy function. Small clusters were discarded based on a function of the average fitting error. In the comparison of different algorithms, RCMSA performed better than PEARL. However, their scalar segmentation error may not fully justify the correctness of the conclusion. 
The data containing a larger number of inliers can  tolerate more outliers for the same segmentation error.
The model complexity was used as one of the parameters and had to be changed according to the specific estimation problem. It was 100 for the fundamental matrix and 10 for homography. Applying the values vice-versa, the estimator will no longer work all the time. This kind of adjustment cannot be done without prior knowledge.

Most of the methods introduced above either used automatically a 2-3 pixels scale assumption for RANSAC, or tried to avoid the scale threshold  but introduced additional parameter(s) for a specific task.  If no empirical values of the parameters are known before the estimation, many experiments are required in order to tune them correctly.
In general,  there is no systematic way to predict the values of these parameters. The scale estimate problem can be solved only if  for each structure separately, the inlier scale is estimated adaptively from the input data.

The generalized projection-based M-estimator (gpbM)  \cite{mittalanand12} solved the robust estimation problem for each iteration in three independent steps: scale estimation, mean shift based recovery of the structure and inlier/outlier dichotomy.  The only parameter to be specified by the user was the number of trials for random sampling, which is required in any robust estimator using elemental subsets.
In the first step, the scale was estimated, with all remaining data involved, by locating the highest value of the cumulative distribution function weighted by its size. This implementation made it easier a structure containing more points  to be detected first.
The weights were critical to obtain the correct scale estimate, as Figure 3 from \cite{mittalanand12} showed. This weighting strategy may not work all the time due to the interaction between inliers and outliers.

In this paper, we propose a new robust estimator for multiple inlier structures which also uses three independent steps: scale estimation, structure recovery, strength based structure classification, but
each step has a completely different implementation from that in gpbM.
The major innovations of the paper are summarized below.
\begin{itemize}
\item The scale estimation is carried out on a small set consisting of points in a single structure.
\item The simplest mean shift algorithm is used.
\item All the input data are classified into different structures first, without using any threshold.
\item Structures are characterized by their strength (average density)
and in general an inlier structure has a stronger strength.
\item The limitations of the new robust estimator are discussed in detail.
\end{itemize}
Our experiments show successful results in different estimation problems, without tuning any parameters. In spite of various objective functions, the entire process is self-adaptive.

The linearized objective functions are presented in Section \ref{sec:prepare}. The algorithm of the new robust estimator is detailed in Section \ref{sec:newscale}.
Experiments of different estimation problems
are given in Section \ref{sec:experiments}.
Finally, in Section \ref{sec:discussion} we discuss some open problems.

\section{From a nonlinear to a linear space}
\label{sec:prepare}

In Section \ref{sec:transformation},
the nonlinear objective function of the inputs is 
transformed into a linear function of the carrier vectors.
The first order approximation of the covariance matrix
of these carriers is also computed.
Section \ref{sec:reducedistance} explains how 
the largest Mahalanobis distance of each input point
is taken into account when 
multiple carrier vectors are derived.

\subsection{Carrier vectors}
\label{sec:transformation}

The nonlinear objective functions in computer vision can be transformed 
into linear relations in higher dimensions.
These linear relations, containing terms formed by the  input measurements and their pairwise products, are called {\it carriers}.
Each relation gives a carrier vector.
For linear objective functions, such as plane fitting, 
the input variables and the carrier vector are identical.

In the estimation of fundamental matrix,
the input data $\by$ are the point correspondences from two images 
$\left[ x\;\;y\;\;x^\prime\;\;y^\prime \right]^\top \in \mathbb{R}^4$, 
with $l=4$ dimensions. 
The objective function with noisy image coordinates is
\begin{equation}
\label{eqn:epipolar}
[x^\prime~~ y^\prime~~ 1] ~\bF~ [x~~ y~~ 1]^\top \simeq  0
\end{equation}
which gives a carrier vector $\bx \in \mathbb{R}^8$ 
containing $m=8$ carriers
\begin{equation}
\label{eqn:epipolarcarrier}
\bx=\left[x\;\;y\;\;x^\prime\;\;y^\prime\;\;x x^\prime\;\;
x y^\prime\;\; y x^\prime \;\;y y^\prime \right]^\top .
\end{equation}
The linearized function of the carriers is
\begin{equation}
\label{eqn:linear}
\bx_{i}^\top \btheta - \alpha \simeq 0 ~~~ i = 1,\ldots,n_{in}
\end{equation}
where $n_{in}$ denotes the number of inliers.
Vector $\btheta \in \mathbb{R}^8$ 
and scalar intercept $\alpha$ 
derived from the $3\times 3$ matrix $\bF$ are to be estimated.
The constraint $\btheta^{\top} \btheta = 1$ eliminates the multiplicative 
ambiguity in (\ref{eqn:linear}). 

In the general case, several linear equations can be derived 
from a single input $\by_i$
\begin{equation}
\label{eqn:linearSolve}
\bx_{i}^{[c]\top} \btheta - \alpha \simeq 0 ~~~ 
c = 1,\ldots,\zeta ~~~ i = 1,\ldots,n_{in} 
\end{equation}
corresponding to $\zeta$ different carrier vectors $\bx^{[c]}$.
For example, the estimation of homography 
has $\zeta =2$ carrier vectors derived from
$x$ and $y$ image coordinates.

The Jacobian matrix is required for the first order approximation of the covariance of carrier vector.
From each carrier vector $\bx^{[c]}$, an $m\times l$ Jacobian matrix
$\bJ_{\scriptsize{\bx^{[c]}|\by}}$ is derived.
Each column of the Jacobian matrix contains the derivatives of the 
$m$ carriers in $\bx^{[c]}$ with respect to one measurement from $\by$.
The Jacobian matrices derived from linear objective
functions are not input dependent, 
while those derived from
nonlinear objective functions rely on the specific input point. 
The carrier vectors are {\it heteroscedastic} for nonlinear objective 
functions. For example, the transpose of the $8\times 4$ Jacobian matrix
of the fundamental matrix
\begin{equation}
\label{eqn:fundamcov}
\bJ_{\scriptsize{\bx_i|\by_i}}^\top = \left[
\begin{array}{@{\hspace{-0.03cm}}c@{\hspace{0.15cm}}c@
{\hspace{0.15cm}} c@{\hspace{0.15cm}}c@{\hspace{0.15cm}}c@
{\hspace{0.15cm}}c@{\hspace{0.15cm}}c@{\hspace{0.15cm}}c@
{\hspace{-0.03cm}}}
1 & 0 & 0 & 0 & x_i^\prime & y_i^\prime & 0 & 0 \\ 
0 & 1 & 0 & 0 & 0 & 0 & x_i^\prime & y_i^\prime \\ 
0 & 0 & 1 & 0 & x_{i} & 0 & y_{i} & 0 \\ 
0 & 0 & 0 & 1 & 0 & x_{i} & 0 & y_{i}  
\end{array} \right] 
\end{equation}
depends on $\by_i$. 

The $l\times l$ covariance matrix of the 
measurements $\sigma^2\bC_{\scriptsize{\by}}$ 
with $\det \bC_{\scriptsize{\by}} = 1$, 
has to be provided before estimation.
This is a chicken-egg problem, since the input points 
have not yet been classified into inliers and outliers.
A reasonable assumption is to set $\bC_{\scriptsize{\by}}$ as the identity 
matrix $\bI_{\scriptsize \by}$, if no additional information is given.
The input data are considered as independent and identically distributed, and contain
{\it homoscedastic} measurements with the same covariance.

The covariance of the carrier vector $\sigma^2 \bC_i^{[c]}$ is
computed from
\begin{equation}
\label{eqn:newcovariance}
\sigma^2\bC_i^{[c]} = \sigma^2 \bJ_{\scriptsize{\bx_i^{[c]}|
\by_i}} ~\bC_{\scriptsize{\by}} ~ \bJ_{\scriptsize{\bx_i^{[c]}|
\by_i}}^\top 
\end{equation}
where the dimensions of $\bC_i^{[c]}$ is $m\times m$.
The scale $\sigma$ of the structure is unknown and to be estimated.

\subsection{Computation of the Mahalanobis distances}
\label{sec:reducedistance}

The elemental subset needs $m_e = \lceil \frac{m}{\zeta} \rceil$ 
input points to uniquely define $\btheta$ and $\alpha$ in the linear space.
For example, the homography $(m = 8,\; \zeta = 2)$ requires four point 
pairs, and eight pairs are necessary for 
the fundamental matrix $(m = 8,\; \zeta = 1)$ 
if the 8-point algorithm is used.
The input data should be normalized \cite[Section 4.4]{hartley04},
and the obtained structures mapped back onto the original space.

For each $\btheta$, 
every carrier vector is projected to a scalar value
$z_i^{[c]} = \bx_i^{[c]\top} \btheta ,\; c=1,\ldots,\zeta$.
The average projection of the $m$ vectors
from an elemental subset is $\alpha$.
The variance of $z_i^{[c]}$ is 
$\sigma^2 H_i^{[c]} = \sigma^2\btheta^\top \bC_i^{[c]} \btheta$.

The Mahalanobis distance, scaled by an unknown $\sigma$,  indicating how far is a projection $z_i^{[c]}$ 
from $\alpha$, is computed from
\begin{eqnarray}
\label{eqn:distances} \nonumber
d_i^{[c]} &=& \sqrt{\left( \bx_i^{[c]\top} \btheta  -
\alpha \right)^\top \left(H_i^{{[c]}}\right)^{-1} \left( 
\bx_i^{[c]\top} \btheta - \alpha \right)} \\
&=& \frac{|\bx_i^{[c]\top} \btheta  - \alpha |}
{\sqrt{\btheta^\top \bC_i^{[c]} \btheta}} \qquad c=1,\ldots,\zeta .
\end{eqnarray} 
Each input point $\by_i$ gives a 
$\zeta$-dimensional Mahalanobis distance vector
\begin{equation}
\label{eqn:mahavector}
\bd_i = \left[ \, d_i^{[1]} ~\ldots ~d_i^{[\zeta]} \, \right]^\top
\qquad i=1,\ldots, n .
\end{equation}
The worst-case scenario is taken to retain the largest Mahalanobis 
distance $d_i^{[\tilde{c}_i]}$ from all the $\zeta$ values
\begin{equation}
\label{eqn:jmax}
\tilde{c}_i = \operatorname* {arg\,max}_{c=1,\ldots,\zeta} \; d_i^{[c]}.
\end{equation}
For different $\btheta$-s, the same input point may 
have its largest distance computed from different carriers.

The symbols related to the largest Mahalanobis distance are:
$\tilde{d}_i$, the largest Mahalanobis distance for input $\by_i$;
$\tilde{\bx}_i$, the corresponding $m\times 1$ carrier vector;
$\tilde{z}_i$, the scalar projection of $\tilde{\bx}_i$;
$\widetilde{\bC}_i$, the $m\times m$ covariance matrix of $\tilde{\bx}_i$;
$\widetilde{H}_i$, the variance of $\tilde{z}_i$; 
$\hat\sigma$, the scale multiplying $\widetilde{\bC}_i$ 
and $\widetilde{H}_i$, which has to be estimated.

\section{Estimation of Multiple Structures} 
\label{sec:newscale}

The new algorithm is detailed in this section. The scale $\hat\sigma$ 
for a structure is estimated in Section \ref{sec:newsigma} by an
expansion criteria. The estimated scale is used in the mean shift 
to re-estimate the structure in Section \ref{sec:newstructure}. 
The iterative process continues until not enough 
input points remain
for a further estimation. In Section \ref{sec:strength}, 
all the estimated structures are ordered by strengths with  
the strongest inlier structures returned first.
The limitations of the method are explained in 
Section \ref{sec:inlieroutlier}.

\subsection{Scale estimation for a structure}
\label{sec:newsigma}

Assume that $n$ input points remain in the current iteration.
The estimation process starts with $M$ randomly
generated elemental subsets, each giving a $\btheta$ and 
an $\alpha$.

For every point $\by_i$, compute the largest Mahalanobis distance 
$\tilde{d}_i$
\begin{equation}
\label{eqn:orderdistances}
\tilde{d}_i = \frac{|\tilde{\bx}_i^\top \btheta  -
\alpha |}{\sqrt{\btheta^\top \widetilde{\bC}_i \btheta}} \ge 0 \qquad 
i=1,\ldots,n.
\end{equation} 
Sort the $n$-distances in ascending order, denoted 
$\tilde{d}_{[i]}$. 
In total $j=1,\ldots,M$ sorted sequences $\tilde{d}_{[i,j]}$ are found from all the trials.

Let $n_{\epsilon} \ll n$ represent a small amount of points
\begin{equation}
\label{eqn:ntotal}
n_{\epsilon} = \frac{\epsilon \, n}{100} \qquad 0 < \epsilon \ll 100
\end{equation}
where $\epsilon$ defines the size of $n_{\epsilon}$ in percentage
of the input amount.

Among all $M$ trials, find the sequence 
that gives the {\it minimum sum} of Mahalanobis distances
from the first $n_{\epsilon}$ points
\begin{equation}
\label{eqn:minmaha}
\min_{M} \sum_{i=1}^{n_{\epsilon}} \tilde{d}_{[i,j]}.
\end{equation}
This sequence is denoted as $\tilde{d}_{[i]_M}$
and contains $n$ points in total.
The first $\tilde{n}_{\epsilon}$ points are collected as the initial set. 

If inlier structures still exist and $M$ is sufficiently large, these $\tilde{n}_{\epsilon}$ points have a high probability to be selected
from a single inlier structure,
since it is more dense than the outliers. 
Neither information on the number of structures, 
nor the inlier amounts for each structure is known beforehand
to establish $\tilde{n}_{\epsilon}$ deterministically. 

Two rules should be considered for the ratio $\epsilon\%$. 
First, $\tilde{n}_{\epsilon}$ should be smaller than the size of any inlier
structure to be estimated. Therefore, a small ratio is 
preferred to detect potential structures.
In the following sections, all our experiments start with 
$\epsilon\% = 5\%$.
The second rule is to have the size of $\tilde{n}_{\epsilon}$ at least five
times the number of points in the elemental subset,
as suggested in \cite[page 182]{hartley04}.
This condition reduces unstable results when relatively
few input points are provided.

\begin{figure}[t]
\centering
\begin{tabular}{@{\hspace{-0.0cm}}c@{\hspace{-0.0cm}}c}
\includegraphics[scale=0.3]{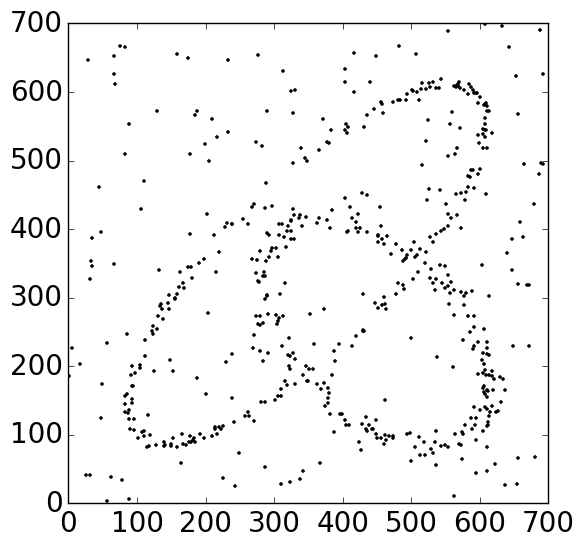}&
\includegraphics[scale=0.3]{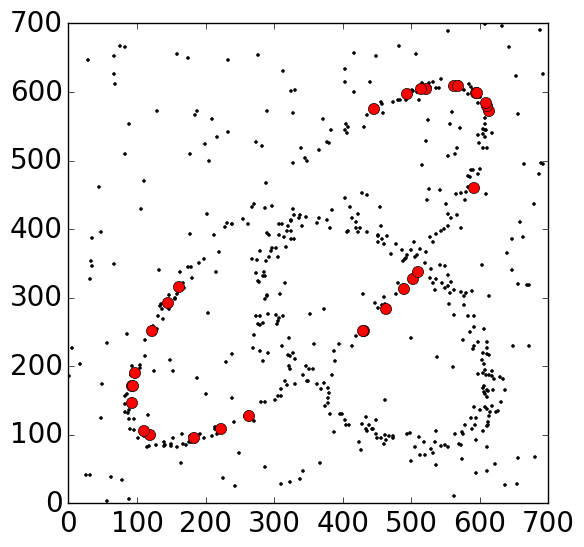} \\
(a) & (b) \\ 
\includegraphics[scale=0.23]{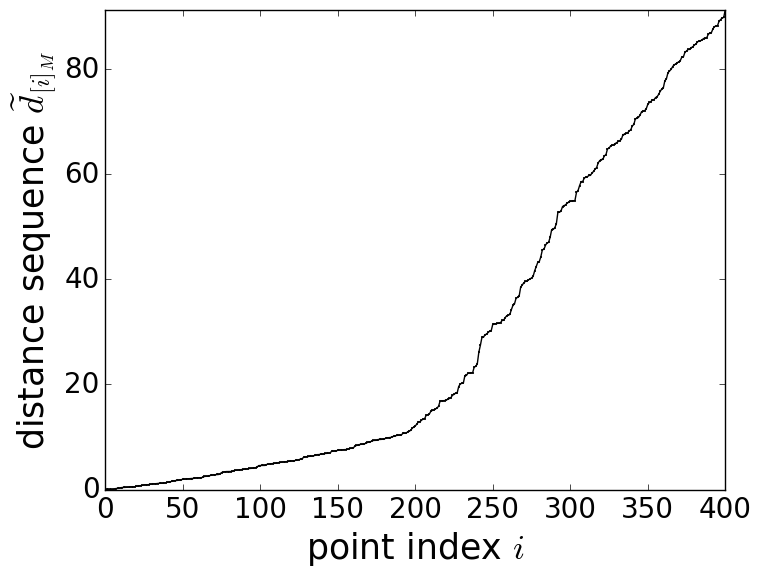} &
\includegraphics[scale=0.23]{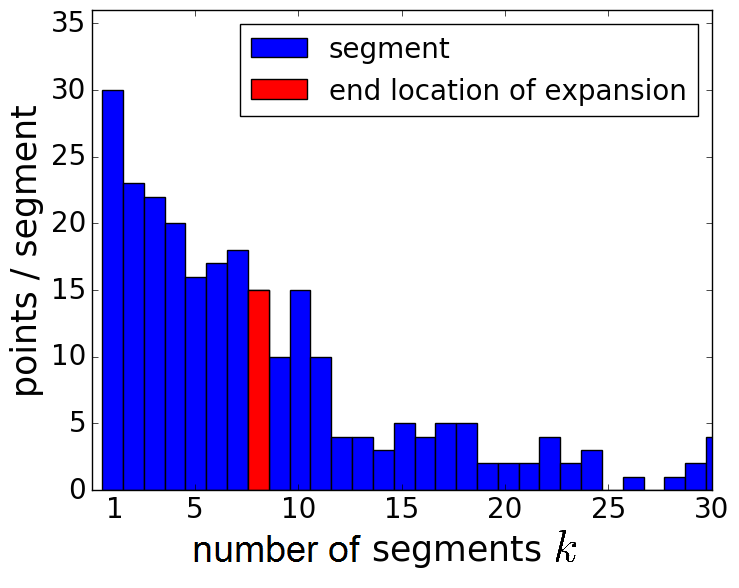} \\
(c) & (d) \\
\includegraphics[scale=0.23]{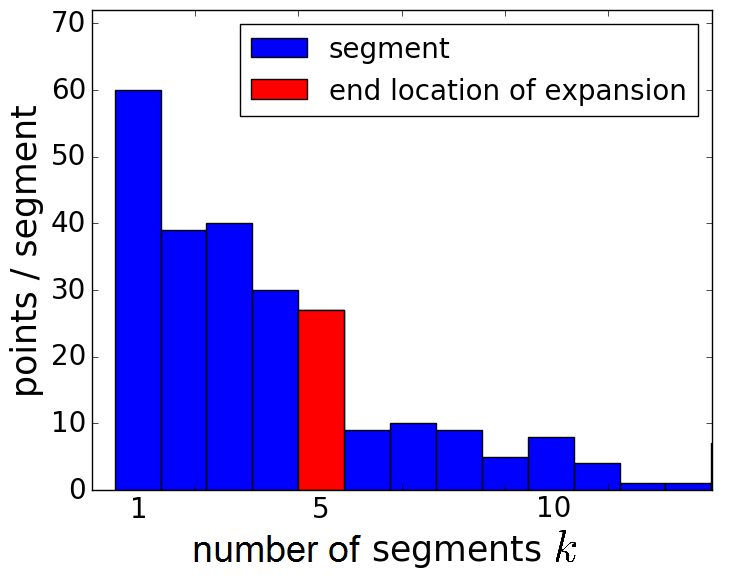} &
\includegraphics[scale=0.23]{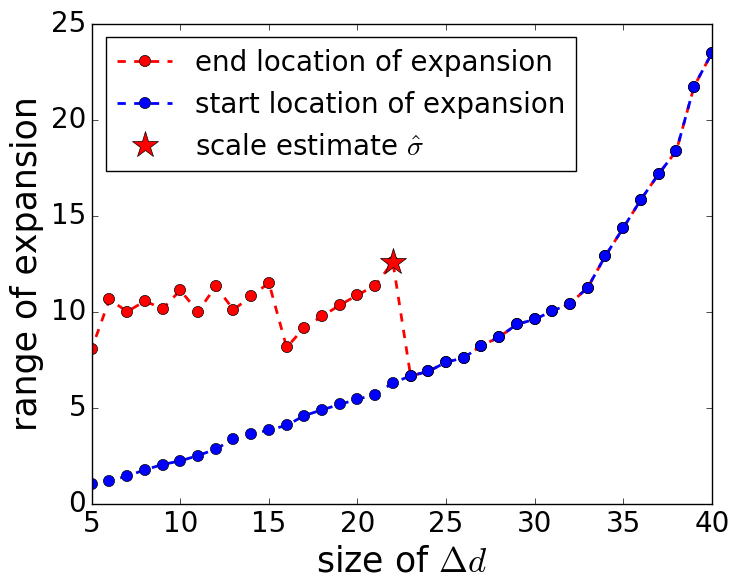} \\
(e) & (f)
\end{tabular}
\caption{Scale estimation.
(a) Input data.
(b) Initial set for an iteration.
(c) The first 400 points in the sequence $\tilde{d}_{[i]_M}$.
(d) Histogram of point amounts with segment size 
$\Delta d_0 = \tilde{d}_{[5\%]_M}$.
(e) Histogram of point amounts with segment size 
$\Delta d_5 = \tilde{d}_{[10\%]_M}$.
(f) Expansion criteria applied to increasing sets.}
\label{fig:expand}
\end{figure}

To recover the scale $\hat \sigma$, as many  possible
points belonging to the same structure
have to be classified together.
The following example justifies the use of
expansion criteria for scale estimation.

In Fig.\ref{fig:expand}a two ellipses are shown; 
each of them has $n_{in} = 200$ inlier points.
They are corrupted by 
different Gaussian noise with standard deviation 
$\sigma_g = 5$ and $10$ respectively. Another
$n_{out} = 200$ outliers are randomly placed in the $700\times 700$ image.
With $M = 2000$ and $\epsilon\% = 5\%$
($\tilde{n}_{\epsilon} = 30$) we obtain the initial set in Fig.\ref{fig:expand}b.
The distances in $\tilde{d}_{[i]_M}$ for the first 400 points from 600 in total, are shown in Fig.\ref{fig:expand}c.

Divide the sequence $\tilde{d}_{[i]_M}$ into multiple segments,
and each segment has an equal range of Mahalanobis distance, 
$\Delta d$.
Let $n_{k}$ denote the number of points within 
the $k$-th segment, $k = 0, 1, 2, \ldots$. The $k=0$ is the start
$\tilde{n}_{\epsilon} = n_0$ and the average point density in this
segment equals to $n_0$.
The expansion process verifies the following condition 
for each $k$
\begin{equation}
\label{eqn:condition} 
\frac{n_{k+1}}{\frac{1}{k}\sum_{i = 1}^{k}{n_i}} \leq 0.5
\end{equation}
where the numerator is the number of points
in the ($k+1$)-th segment and the denominator is the average point numbers inside all the $k$ segments. 
When the point density drops below half of 
the average in the previous segments, 
the boundary to separate this structure
from the outliers is found, $k=k_t$.
The value of the scale estimate is $k_{t}\Delta d$.

Due to the randomness of the input data,
a single estimation of the scale is not  enough. 
If the size of initial set is too small compared with the true structure, the scale estimate can also be too small to
fully recover the complete structure in the mean shift.

In Fig.\ref{fig:expand}d the expansion 
starting with $\Delta d_0= \tilde{d}_{[5\%]_M}$, the initial set, 
and stops at $k_{t_0}=8$, 
giving $\hat\sigma = 8.06$ (red bar).
This is a relatively small estimate
since a scale larger than 10 is expected 
when $\sigma_g = 5$.
In Fig.\ref{fig:expand}e the sampling with a larger $\Delta d_5 = \tilde{d}_{[10\%]_M}$ 
has its expansion stopped at $k_{t_5}=5$ 
giving $\hat\sigma = 11.10$.
This shows that various estimates can be generated
from different segments layouts, 
and it is similar to the discretization effect over the scale space in
SIFT \cite{lowe04}.

In Fig.\ref{fig:expand}f the expansion process
is applied to an increasing sequence of sets.
The $\Delta d$ starts from 
$\tilde{d}_{[(\epsilon + j)\%]_M}$, $j = 0, 1, 2, \ldots$,
and increases by $1\%$ for each new sampling.
The blue points in the figure
indicate the length of $\Delta d$ in percentage 
of points used as the segment size. 
Every expansion process is performed separately,
and stops at the corresponding red point 
when condition (\ref{eqn:condition}) is met.
The length of $\Delta d$ continues to increase until
it reaches the bound, $j=T+1$, where the sets can no longer expand,
as it is $23\%$ in Fig.\ref{fig:expand}f. 

The scale estimate is found from this {\it region of interest}
where the sets of points can expand.
In Fig.\ref{fig:expand}f it ranges from $5\%$ to $22\%$.
The expansion process may not always be able to start from 
$\Delta d = \tilde{d}_{[\epsilon\%]_M}$, 
but stops immediately at $k = 1$.
Then as $\Delta d$ increases, 
the starting point of region of interest is 
at the place where the expansion process begins. 

The largest estimate from the region of interest
gives the scale $\hat\sigma$
\begin{equation}
\label{eqn:estimatescale}
\hat\sigma = \max_{j=0,\ldots,T}  k_{t_j} \Delta d_j 
\qquad 
\mbox{in region of interest}
\end{equation}
the farthest expansion inside the region of interest. 
In Fig.\ref{fig:expand}f the scale estimate is $\hat\sigma = 12.54$.
From the sequence $\tilde{d}_{[i]_M}$, collect
all the points within the scale estimate for the next step.

If the scale estimator locates an outlier structure,
$\hat\sigma$ in general is much larger and
the structure has weaker strength than inlier
structures, as will be discussed in Section \ref{sec:strength}.
The condition (\ref{eqn:condition}) is a heuristic criteria
since the true distribution of the inliers is unknown.
However, it does not play a sensitive role in the estimation process,
as will be shown in Section \ref{sec:experiments}.
Some methods mentioned in the introduction assumed a Gaussian 
distribution, or proposed a sophisticated theoretical model to classify inliers. 
These approximations may only be valid in specific problems.

\subsection{Mean shift based structure recovery}
\label{sec:newstructure}

From the points collected in the first step, another
$N \ll M$ elemental subsets are generated. 
Most points in this set come from the same structure 
thus $N = M/10$ is enough.

For each trial all the input points 
are projected by $\btheta$ to a one-dimensional space
$\tilde{z}_i = \tilde{\bx}_i^\top \btheta ,\; i=1,\ldots,n$.
The mean shift  \cite{comaniciu02} moves the
$z$ from $z = \alpha$ to the {\it closest mode} 
\begin{eqnarray}
\label{eqn:kde}
\nonumber \left[\widehat{\btheta},\widehat{\alpha}\right] 
&=& \operatorname*{arg\,max}_{\mathbf{\btheta},\alpha}
\frac{1}{n \hat\sigma} \sum_{i=1}^{n} \kappa \left(\left( z - 
\widetilde{z}_i \right)^\top\widetilde{B}_i^{-1} \left( z - \widetilde{z}_i
\right) \right) \\
&=& \frac{1}{n \hat\sigma}
\operatorname*{arg\,max}_{\mathbf{\btheta}} \left( 
\operatorname*{arg\,max}_{z} f_{\btheta}(z) \right) .
\end{eqnarray}
The variance $\tilde{B}_i$ is computed from
\begin{eqnarray}
\label{eqn:oldfull}
\widetilde{B}_i &=& \hat{\sigma}^2 \widetilde{H}_i = 
\hat{\sigma}^2 {\btheta}^\top {\tilde\bC}_i {\btheta} 
\nonumber\\
&=& \hat{\sigma}^2 {\btheta}^\top \bJ_{\scriptsize{\tilde{\bx}_i|
\by_i}} \bJ_{\scriptsize{\tilde{\bx}_i|\by_i}}^\top {\btheta} 
\end{eqnarray}
with $\bC_{\by} = \bI_{\by}$.

The function $\kappa(u)$ is the profile of a 
radial-symmetric kernel $K(u^2)$ defined only for $u \geq 0$. 
For the Epanechnikov kernel 
\begin{equation}
\label{eqn:profile}
\kappa(u) = \left\{
\begin{array}{rcr}
1 - u &  & \left( z - \widetilde{z}_i \right)^\top\widetilde{B}_i^{-1} 
\left( z - \widetilde{z}_i \right) \leq 1 \  \\
0 &  & \left( z - \widetilde{z}_i \right)^\top\widetilde{B}_i^{-1} 
\left( z - \widetilde{z}_i \right) > 1 .\\
\end{array} \right.
\end{equation}
Let $g(u) = -\kappa'(u)$ and for the Epanechnikov kernel, $g(u) = 1$ when
$0 \le u \leq 1$ and $0$ if $u > 1$.
All the points inside the window contribute equally in the mean shift. 
The convergence to the closest mode is obtained by assigning zero to the 
gradient of (\ref{eqn:kde}) in each iteration.
The $z_{new}$ is updated from the current value $z = z_{old}$ by
\begin{eqnarray}
\label{eqn:origmean}
z_{new}= \left[\sum_{i=1}^{n} {g\left(u\right)} \right]^{-1} \!\!
\left[\sum_{i=1}^{n}
{g\left(u\right)} \widetilde{z}_i \right] .
\end{eqnarray}
Many of the $n$ input points have their projections 
more distant from $z_{old}$ than $\pm\widetilde{B}_i$ and their 
weights are zeros.

The highest mode among all $N$ trials gives the estimate 
$\hat z = \hat\alpha$. The vector $\hat\btheta$ is obtained
from the same elemental subset which gives the highest mode.
All the input points that can converge into the
$\pm \hat\sigma$ region around $\hat\alpha$ are 
classified as inliers, resulting in $n_{in}$ points. 
The total least squares (TLS) estimate for the structure is computed
to obtain $\hat\btheta^{tls}$, 
$\hat\alpha^{tls}$ and ${\hat\sigma}^{tls}$.

\subsection{Strength based classification}
\label{sec:strength}

After the mean shift step, the $n_{in}$ points 
are removed from the inputs before the next iteration. 
If the remaining data are not enough for another initial set,
the algorithm terminates and all the recovered structures 
are sorted by their strengths in descending order.
The {\it strength of a structure} is defined as
\begin{equation}
\label{eqn:strength}
s = \frac{n_{in}}{\hat \sigma^{tls}}.
\end{equation}
which can also be seen as the density in the linear space of that structure. 
The value $n_{in}$ represents the point amount 
removed at each iteration and it can be either a
structure of inliers or outliers. 

Structures with stronger strengths are detected first,
and in general are inlier structures with more dense points
and smaller scales.
The new method does not rely on a threshold to separate inliers from the outliers.
After the input data are segmented,
the difference between the structures declared as inliers with stronger strengths and the first outlier structure is clear. 
With the results sorted by strength,
the user has an easy task to retain the inlier structures,
as the examples in Section \ref{sec:experiments} will show.
If an ambiguous inlier/outlier threshold appears, like in Fig.\ref{fig:synthellipse}d,
the strongest inlier structures are still detected correctly.

\subsection{Limitations}
\label{sec:inlieroutlier}

The major limitation of every robust estimator
comes from the interactions between inliers and outliers.
As the outlier amount increases, eventually the inliers and outliers become less separable in the input space.
In our algorithm most of the processing is done in a linear space,
but the limitation introduced by outliers still exists.
We will illustrate it in the following example.

\begin{figure}[t]
\centering
\begin{tabular}{@{\hspace{-0.0cm}}c@{\hspace{-0.0cm}}c}
\includegraphics[scale=0.3]{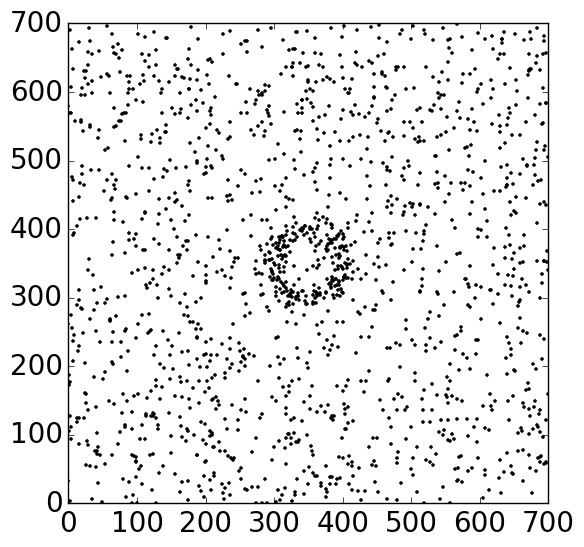}&
\includegraphics[scale=0.3]{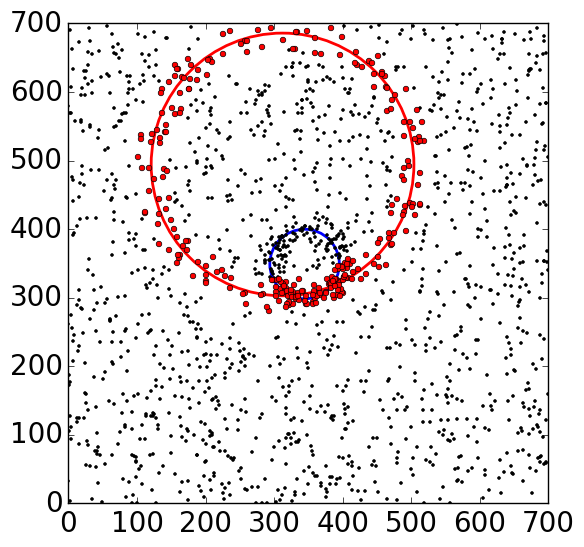} \\
(a) & (b) \\
\includegraphics[scale=0.3]{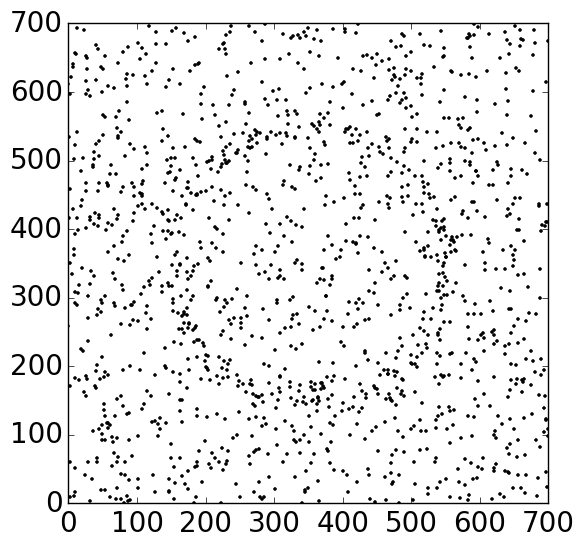}&
\includegraphics[scale=0.3]{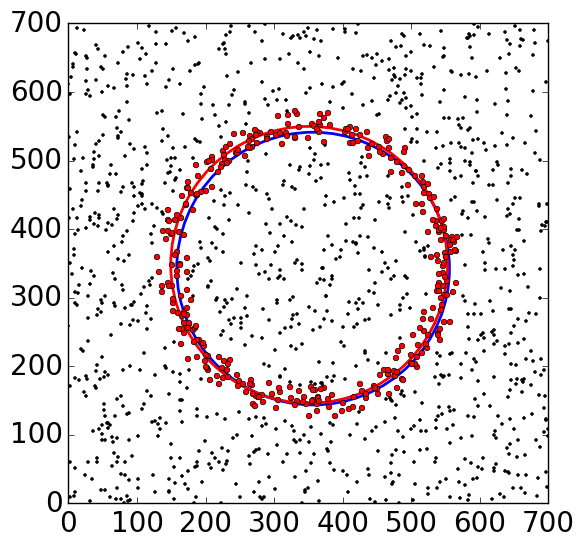} \\
(c) & (d)
\end{tabular}
\caption{The inlier/outlier interaction.
(a) Input data of a circle with radius 50. 
(b) Incorrect final result obtained 
from a correct scale estimate.
(c) Input data of a circle with radius 200. 
(d) Good final result obtained 
from a correct scale estimate.}
\label{fig:outlierproblem}
\end{figure}

In a $700\times 700$ image, a circle consists of $n_{in} = 200$ inliers is
corrupted by Gaussian noise with $\sigma_g = 10$, together with 
$n_{out} = 1500$ outliers.
The first circle has a radius of 50 (Fig.\ref{fig:outlierproblem}a), 
and the other one has a radius of 200 (Fig.\ref{fig:outlierproblem}c).
In both these figures,
the estimator finds the correct scale estimates
from the structure (blue circles) corresponding to the initial sets, where
$\hat\sigma_{50} = 23.65$ and  $\hat\sigma_{200} = 23.58$.

In the true inlier structure,
about 196 points should exist inside the 
scale $\hat\sigma_{50} = 23.65$, 
based on the Gaussian distribution.
The number of outliers in the same location can be roughly estimated as 
\[ \quad (2\pi\, 50) (2*23.65) \frac{1500}{700\times 700}
= 45~ \mbox{points.} \] 
thus about 241 points can be found in the true inlier structure.
However, after the mean shift step an incorrect final result (red circle)
containing 261 points is obtained in Fig.\ref{fig:outlierproblem}b,
where 84 points are true inliers and 177 points from the outliers.
Although the true structure appears more dense in the input space,
the mean shift converges
to an incorrect mode due to the heavy noise from outliers.

The circle in Fig.\ref{fig:outlierproblem}c appears
much weaker, however after 100 tests with randomly generated data 
(inlier/outlier), it returns more stable estimations than the smaller 
circle in Fig.\ref{fig:outlierproblem}a.
In a result shown in Fig.\ref{fig:outlierproblem}d,
346 points are classified as inliers, 
where 190 points are from true inliers
and 156 points from the outliers. The mean shift 
has a much lower probability to converge to another, incorrect mode,  
and this inlier structure resists more outliers.

Similar limitation exists in RANSAC when many outliers 
are present. Even a correct scale given by the user
can still lead to an incorrect estimation. 
The methods proposed in \cite{pham14} and \cite{tennokoon16}
returned incorrect results if too many outliers existed.
The failure of RANSAC also occurs due to not explicitly considering
the underlying task \cite{hassner14}.

In \cite{vedaldi05}, a robust estimator was proposed 
to track objects in an image sequence,
by combining an extended Kalman filter with a structure from motion algorithm.
The Figure 5 in that paper showed that 
the correct estimate was returned with
the data containing more than 60\% outliers.
For the homography estimation in Figure 6 of \cite{moisan12},
more than 90\% of the points were outliers, 
and correct result were obtained after 10000 iterations of
a contrario outlier elimination process. 
Since these results did not provide repetitive tests,
the stability of the methods cannot be verified.

The strength of inlier points
is another factor with a strong influence on
the inlier/outlier interaction.
Firstly, 
the level of the inlier noise affects the number of outliers that can be tolerated.
With the same number of inliers,
structures with lesser inlier noise can be estimated more
robustly since a smaller scale estimate results in a stronger strength. 
The inlier structures with weaker strengths generally have larger noises 
and the scale estimates are also larger.
The outliers will have a stronger interaction with these weak
inlier structures and can lead to spurious results, 
see Fig.\ref{fig:synthellipse}f. 

Secondly, 
when the inlier amount is too small,
the initial set may not closely align with a true structure.
The scale estimation becomes unstable 
since the region of interest can sometimes cover a very narrow range of $\Delta d$ in the expansion process.
The scale estimate could be much smaller than the true value
and only the minority of the points will converge to the inlier structure. 
Instead of a single structure estimate, two or more split structures could be obtained.

\begin{figure}[t]
\centering
\begin{tabular}{@{\hspace{-0.0cm}}c@{\hspace{-0.0cm}}c}
\includegraphics[scale=0.23]{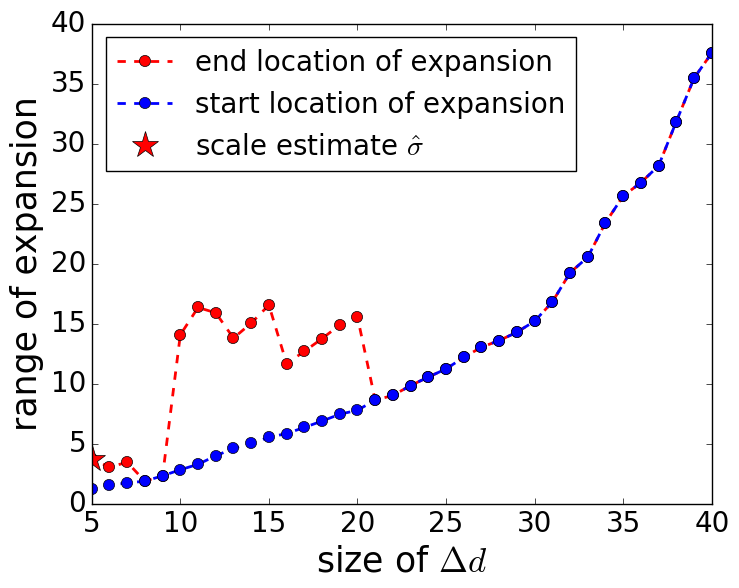}&
\includegraphics[scale=0.23]{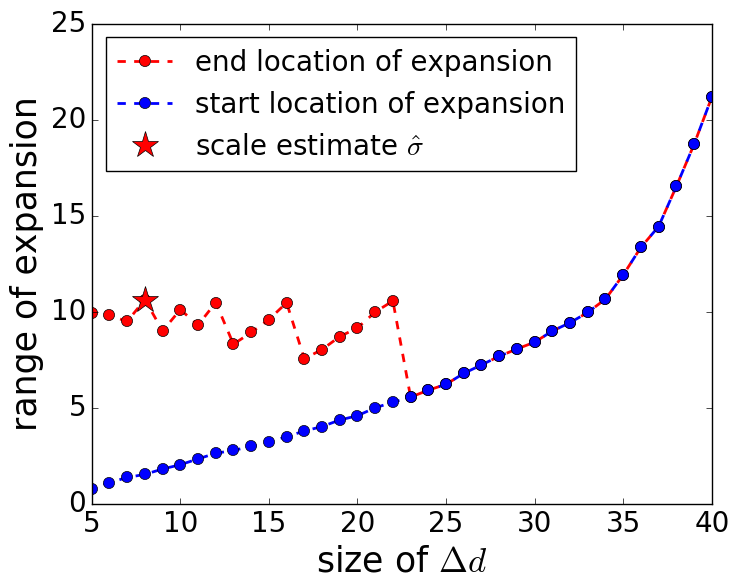} \\
(a) & (b)
\end{tabular}
\caption{Limitation of scale estimation.
(a) Unstable estimate obtained 
from a small number of inliers ($n_{in} = 200, 
n_{out} = 400$). 
(b) More robust estimate obtained 
from a larger number of inliers ($n_{in} = 400, 
n_{out} = 400$).}
\label{fig:limitproblem}
\end{figure}

In Fig.\ref{fig:limitproblem}a,
the expansion process is applied to
the same example as in Fig.\ref{fig:expand}, but
with $n_{out} = 400$.
The expansion stops soon and the algorithm
locates the region of interest between $5\%-7\%$
giving a small scale estimate. 
After applying the expansion criteria
many times,
the range of expansion from different testing data are not stable.
In Fig.\ref{fig:limitproblem}b the number of inliers 
is raised to $n_{in} = 400$.
The scale estimate becomes a more stable value with
the region of interest located between $5\% - 22\%$. 

When the inlier/outlier interaction is strong,
preprocessing on the input data is required to 
obtain more inlier points, and/or reduce the outlier amount
for a better performance.
In Fig.\ref{fig:homographyFig1}a of Section \ref{sec:experiments}, 
an example is given where homography estimation in 2D
is used to segment objects in 3D scene.
Under the small translational motions, the two planes on the bus
though  orthogonal in 3D, are not separable in 2D 
due to the relatively small amount of inlier points.
In Fig.\ref{fig:homographyFig2} we show that by using more inlier points, the estimator will recover more inlier structures.

If an inlier structure appears split in several structures
with fewer points, post-processing is needed to merge them.
The user can easily locate them by their strengths
since most of these split structures are still
stronger compared with the outliers.
The similarity of two structures should be compared
in the input space where measurements are obtained,
as the derived carriers in the linear space
do not represent the nonlinearities of the inputs explicitly.

For two inlier structures with linear objective function, 
the merge can be implemented based on the 
orientation of each structure and the distance between them.
For two ellipses, the geometric tools to determine the overlap area 
can be used \cite{hughes12}.
The measurements of fundamental matrices and the homographies are in the projective space instead of euclidean.
If the reconstructed 3D scene can be provided 
from auto-calibration \cite[Chapter 19]{hartley04},
the 3D information should be applied to separate or merge
the two structures.

\subsection{Review of the algorithm}
\label{sec:flowchart}

The new algorithm is summarized below.

\noindent\hrulefill\\
{\bf \centerline{
Robust estimation of multiple inlier structures}}
{\vspace{0.01cm}
\noindent\hrulefill}\\
\noindent{\bf Input}:
$\by_i$, $i=1,\ldots,n$ data points that contain an unknown number of 
inlier structures with their scales unspecified, along with outliers.
The covariance matrices for $\by_i$ are 
$\bC_{\scriptsize\by}=\bI_{\scriptsize\by}$ 
if not provided explicitly.

\noindent{\bf Output}:
The sorted structures with inliers come out first.

\begin{itemize}
\item Compute the carriers $\bx_i^{[c]}$, $c=1,\ldots,\zeta$, and the 
Jacobians $\bJ_{\scriptsize{\bx_i^{[c]}|\by_i}}$, for each input
$\by_i$,  $i=1,\ldots,n$.

\vspace{0.3cm}
\item[$\odot$] Generate $M$ random trials 
based on elemental subsets. 
\begin{itemize}
\item For each elemental subset find $\btheta$ and $\alpha$.
\item Compute the Mahalanobis distances from $\alpha$ for all carrier
vectors $\bx_i^{[c]}$, $c=1,\ldots,\zeta$. Keep the largest distance 
$\tilde{d}_i$ for each point.
\item Sort the Mahalanobis distances in ascending order.
\item Among all $M$ trials, find the sequence $\tilde{d}_{[i]_M}$ 
with the minimum sum of distances for $n_{\epsilon}$ 
points remained for processing.
\end{itemize}
\item Apply the expansion criteria to an increasing 
sequence of sets and determine
the region of interest for a structure.
\item In the region of interest find the 
largest estimate as $\hat\sigma$ and collect all points inside this scale.
\item Generate $N \ll M$ random trials from these points.
\begin{itemize}
\item Apply the mean shift to all the existing points,
to find the closest mode from $\alpha$.
\item Find $\hat\alpha$ at the maximum mode among all $N$ trials, 
and $\hat\btheta$ from the same elemental subset.
\item The recovered structure contains
$n_{in}$ points which converged to
$\pm \hat\sigma$ from $\hat\alpha$. 
\end{itemize}
\item Compute the TLS solution for the structure and
remove the $n_{in}$ points from the inputs. 
\item Go back to $\odot$ and start another iteration.
\item If not enough input points remain, 
sort all the structures by their strengths and return the result.
\end{itemize}
\noindent\hrulefill

\section{Experiments}
\label{sec:experiments}

Several synthetic and real examples are presented in this section.
In most cases a single carrier vector exists, $\zeta = 1$, 
except for the homography estimation 
which has two and $\zeta = 2$. 
The Epanechnikov kernel is used in the mean shift.

The input data for synthetic problems are generated 
randomly and Gaussian noise is added to each inlier structure.
The standard deviation $\sigma_g$ is specified 
only to verify the results, while not used in the estimation process.
The values of the scales and point amounts for each structure
are returned as the output of the algorithm.
The processing time for an i7-2617M 1.5GHz PC is given.

\subsection{2D Line}
\label{sec:2dline}

\begin{figure}[t]
\centering
\begin{tabular}{@{\hspace{-0.0cm}}c@{\hspace{-0.0cm}}c}
\includegraphics[scale=0.3]{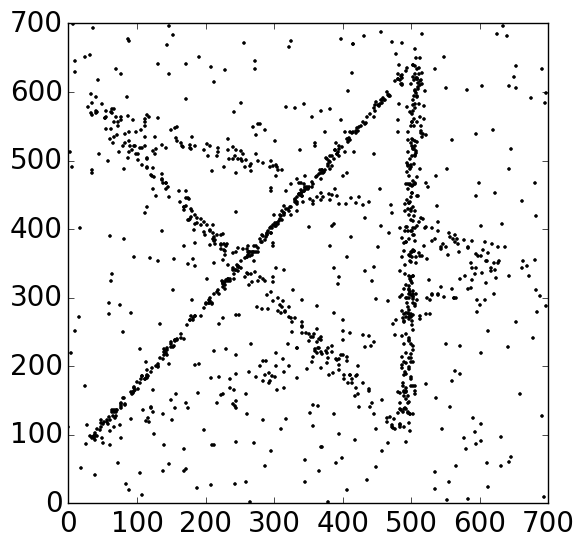}&
\includegraphics[scale=0.3]{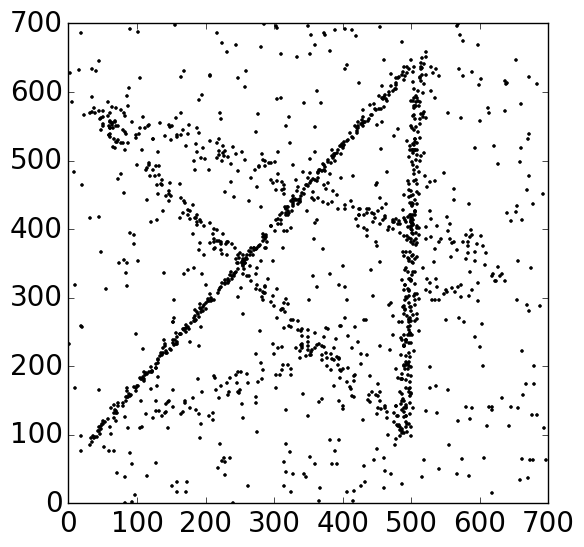} \\
(a) & (b) \\ 
\includegraphics[scale=0.3]{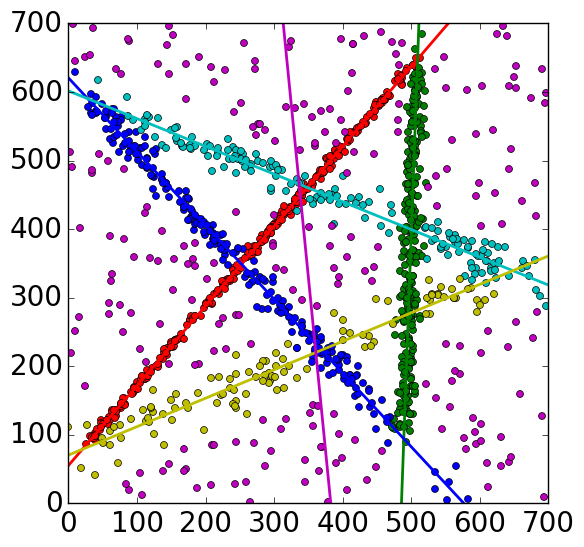}&
\includegraphics[scale=0.3]{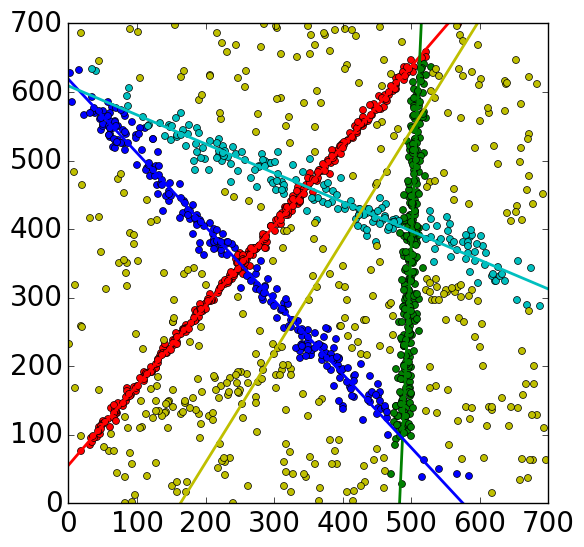}\\   
(c) & (d) \\
\includegraphics[scale=0.3]{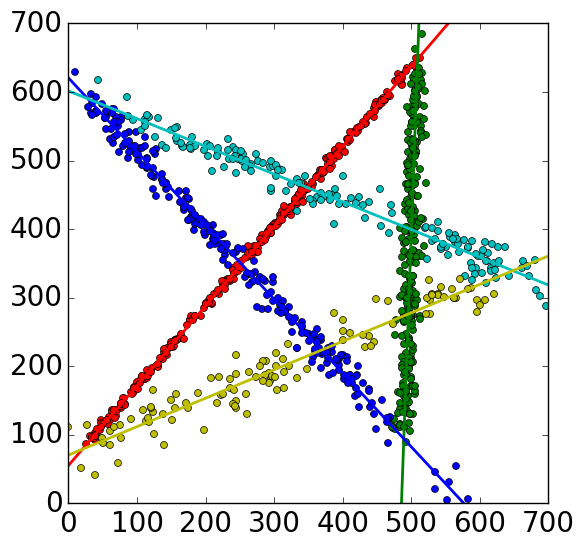}&
\includegraphics[scale=0.3]{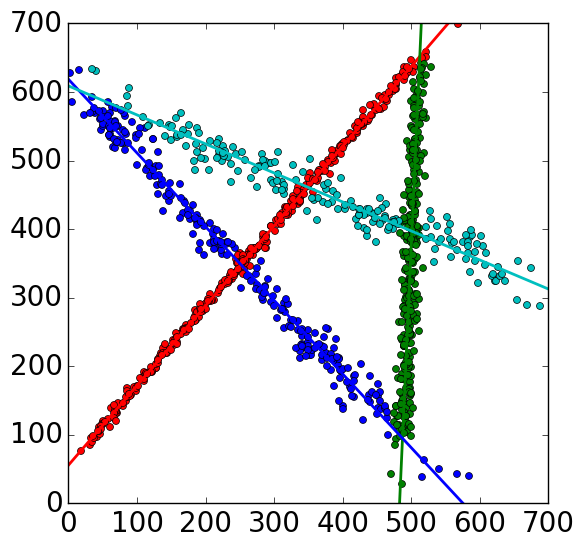}\\   
(e) & (f) 
\end{tabular}
\caption{Synthetic 2D line estimations. 
(a) Case 1: five lines with 350 outliers.  
(b) Case 2: five lines with 500 outliers. 
(c) Recovered six structures, case 1. 
(d) Recovered five structures, case 2.
(e) Five strongest structures, case 1.
(f) Four strongest structures, case 2.}
\label{fig:synthle}
\end{figure}

In the first example {\it multiple 2D lines} are estimated. 
The noisy objective function is
\begin{equation}
\label{eqn:line}
\theta_1 x_{i} + \theta_2 y_{i} - \alpha \simeq 0 \qquad 
i=1,\ldots,n_{in}.
\end{equation}
The input variable $\by = [x ~ y]^\top$ is identical with
the carrier vector $\bx$.

Five lines are placed in a $700\times700$
plane (Fig.\ref{fig:synthle}a) and
corrupted with different two-dimensional Gaussian noise.
They have $n_{in} = 300, 250, 200, 150, 100$ inlier points,
and $\sigma_g = 3, 6, 9, 12, 15$, respectively.
Another 350 unstructured
outliers are uniformly distributed in the image. 
The amount of points inside 
each inlier structure is 
small compared to the entire data.

With $M = 1000$, a test result is shown in Fig.\ref{fig:synthle}c. 
The algorithm recovers six structures
\[\begin{array}{rcccccc}
  & red  & green & blue & cyan & yellow & purple\\
scale: & 9.6  & 18.7 & 28.1 & 37.1 & 44.2 & 370.8\\
inliers: & 321 & 282 & 240 & 161 & 106 & 240\\
strength: & 33.4 & 15.1 & 8.5 & 4.3 & 2.4 & 0.6.
\end{array} \]
The first five structures are inliers with stronger strengths as Fig.\ref{fig:synthle}e shows.
The sixth structure, is formed by outliers distributed over the whole image.

When the randomly generated inputs are tested 
independently for 100 times, 
the first four lines are correctly segmented in all the tests.
In the other six tests  
the weakest line ($n_{in}=100,~ \sigma_g = 15$) is not correctly
located.
Of  the 94 correct estimations,
the average result of the scale estimates and 
the classified inlier amounts as well as their respective  standard deviations are
\[\begin{array}{rccccc}
scale: & 10.48 & 19.94& 29.36 & 36.86 & 38.17\\
& (1.17) & (2.44) & (5.30) & (10.40) & (18.29)\\
inliers: & 335.9 & 285.8 & 240.8 & 155.6 & 93.4\\
& (8.2) & (9.5) & (21.3) & (27.0) & (28.4).\\
\end{array} \]
The average processing time is 0.58 seconds.
The estimated scale covers about  $3\sigma_g$ area of an inlier structure. 
In general, the number of classified inliers is larger than the
true amount due to the presence of outliers in the same area.

As the outlier amount increases, the weakest structure
will gradually blend into the background, and the expansion
criteria can hardly separate inliers and outliers for this structure.
In Fig.\ref{fig:synthle}b a case is shown
where the outlier amount is raised to 500 instead of 350,
and the number of inliers remain the same.
In Fig.\ref{fig:synthle}d five structures are returned
\[\begin{array}{rccccc}
  & red  & green & blue & cyan & yellow\\
scale: & 12.0  & 18.6 & 33.8 & 41.5 & 483.6\\
inliers: & 351 & 304 & 256 & 182 & 407\\
strength: & 29.3 & 16.4 & 7.6 & 4.4 & 0.8.
\end{array} \]
The last structure is mixed with the outliers, and
thus not estimated correctly. 
The first four inlier structures are still retained based 
on the strength (Fig.\ref{fig:synthle}f).
In 100 tests the weakest line is detected 64 times, 
the fourth structure ($n_{in} = 150,~ \sigma_g = 12$) 98 times,
and the three strongest structures are estimated 
correctly in all the trials. 

\begin{figure}[t]
\centering
\begin{tabular}{@{\hspace{-0.0cm}}c@{\hspace{0.1cm}}c}
\includegraphics[scale=0.22]{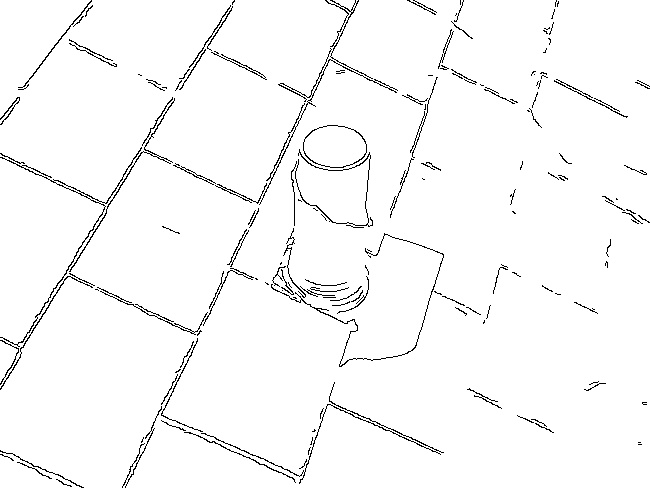}&
\includegraphics[scale=0.215]{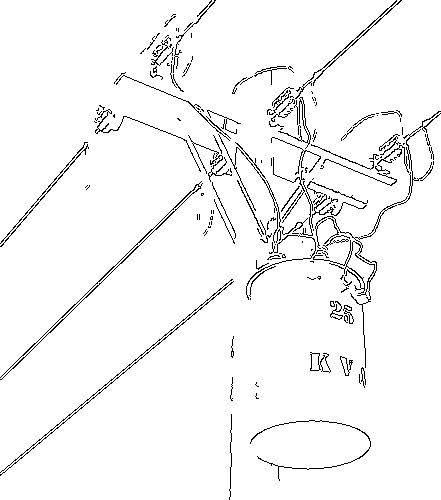}\\   
(a) & (b) \\
\includegraphics[scale=0.22]{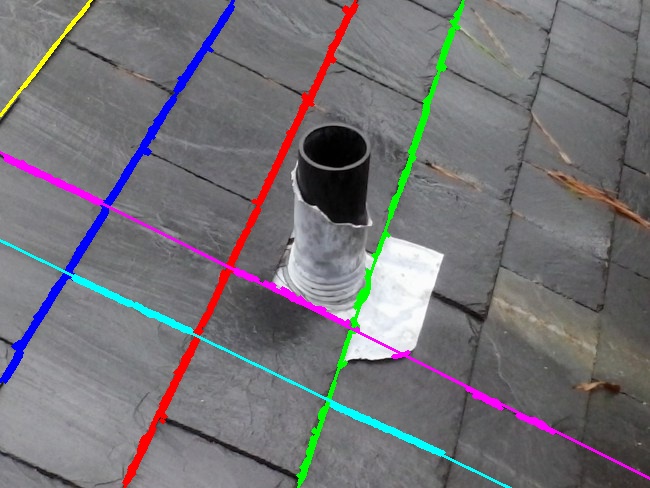}&
\includegraphics[scale=0.215]{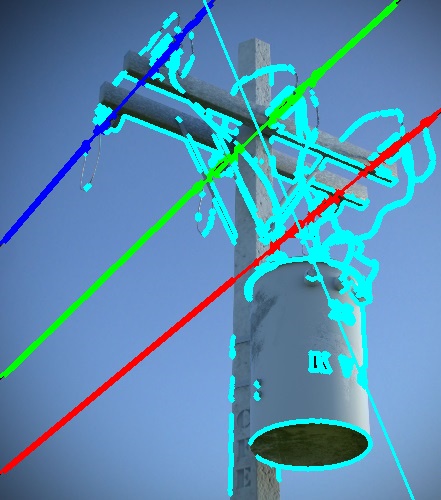}\\   
(c) & (d) 
\end{tabular}
\caption{2D lines in real images. 
(a) {\it Roof}: Canny edges, 8310 points.
(b) {\it Pole}: Canny edges, 8072 points.
(c) {\it Roof}: Six strongest inlier structures.
(d) {\it Pole}: Three strongest inlier structures and one outlier structure.}
\label{fig:realline}
\end{figure}

In Fig.\ref{fig:realline}a and Fig.\ref{fig:realline}b,
the Canny edge detection extracts 
similar sizes of input data (8310 and 8072 points)
from two real images. 
Again with $M=1000$,
the six strongest line structures are 
superimposed over the original image in Fig.\ref{fig:realline}c.
In Fig.\ref{fig:realline}d the three line structures together with the first 
outlier structure are shown.
The processing time depends on the number of structures that detected
by the estimator, 
these two estimations take 7.44  and 4.35 seconds, respectively.

\subsection{2D Ellipse}
\label{sec:2dellipse}

In the next experiment {\it multiple 2D ellipses} are estimated. 
The noisy objective function is
\begin{equation}
\label{eqn:ellipse}
(\by_i - \by_c)^\top \bQ (\by_i - \by_c) - 1 \simeq 0
\qquad i=1,\ldots,n_{in}
\end{equation}
where $\bQ$ is a symmetric $2\times 2$ positive definite matrix and
$\by_c$ is the position of the ellipse center. Given the input variable 
$\by = [x ~ y]^\top$, the carrier is derived as 
$\bx = [x ~ y ~ x^2 ~ xy ~ y^2 ]$. 
The condition $4 \theta_3 \theta_5 - \theta_4^2 > 0$ also
has to be satisfied in order to represent an ellipse.
We also enforce the constraint that 
the major axis cannot be more than 10 times longer than the 
minor axis, to avoid classifying line segment as
a part of a very flat ellipse.

The transpose of the $5\times 2$ Jacobian matrix is
\begin{equation}
\label{eqn:ellipjacob}
\bJ_{\scriptsize{\mathbf{x} | \mathbf{y}}}^\top = 
\left[ \begin{array}{ccccc} 
1 & 0 & 2x & y & 0 \\
0 & 1 & 0  & x & 2y
\end{array} \right] .
\end{equation}
The ellipse fitting is a nonlinear estimation and biased,
especially for the part with large curvature. 
When the inputs are perturbed with zero mean Gaussian noise with 
$\sigma_g$, the standard deviation of carrier vector $\bx$
relative to the true value $\bx_o$ is not zero mean
\begin{equation}
\bE (\bx -\bx_o) = [0 ~~~ 0 ~~~ \sigma^2_g ~~~ 0 ~~~ \sigma^2_g ]^\top
\end{equation}
since the carrier contains $x^2,\;y^2$ terms.
A bias in the estimate can be clearly seen 
when only a small segment of the noisy ellipse is given
in the input. 
Taking into account also the second order statistics 
in estimation still does not  eliminate the bias. 
See  papers \cite{kanatani06},
\cite{szpak15} and their references for additional methods.

\begin{figure}[t]
\centering
\begin{tabular}{@{\hspace{-0.0cm}}c@{\hspace{-0.0cm}}c}
\includegraphics[scale=0.3]{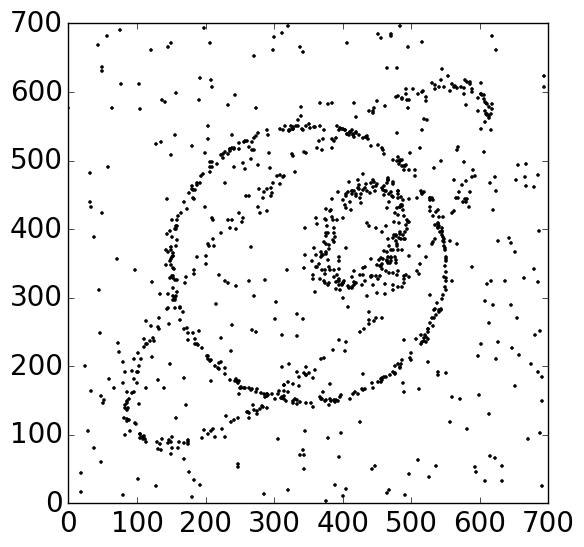}&
\includegraphics[scale=0.3]{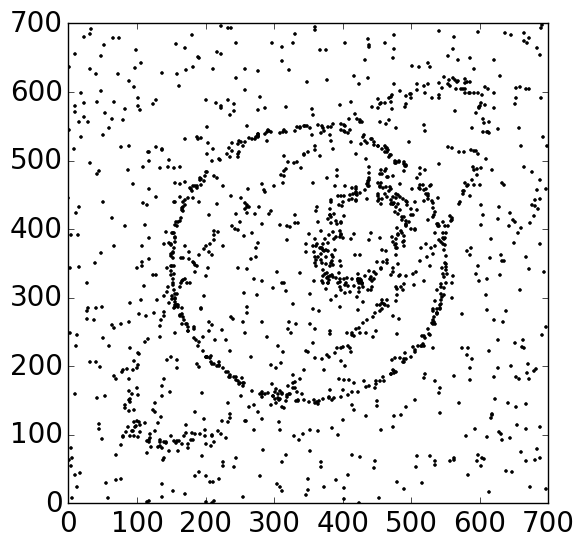}\\   
(a) & (b) \\
\includegraphics[scale=0.3]{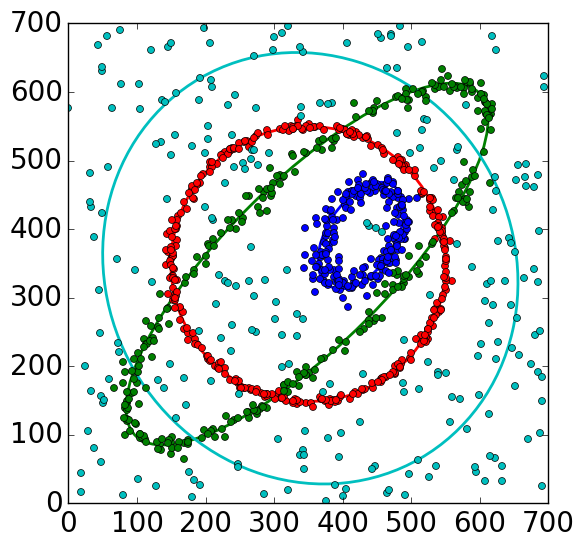}&
\includegraphics[scale=0.3]{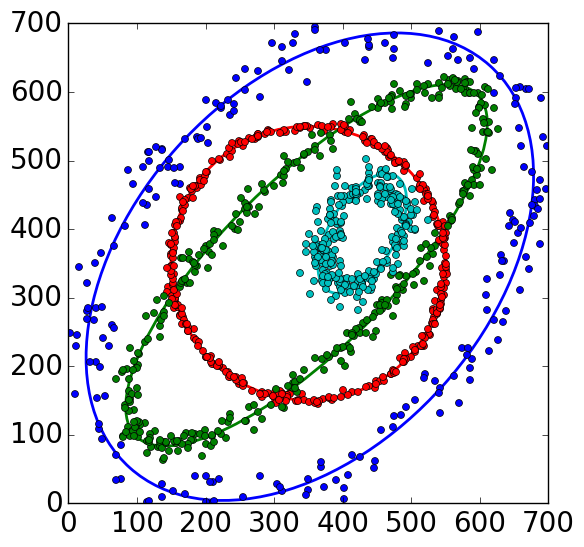}\\   
(c) & (d) \\
\includegraphics[scale=0.3]{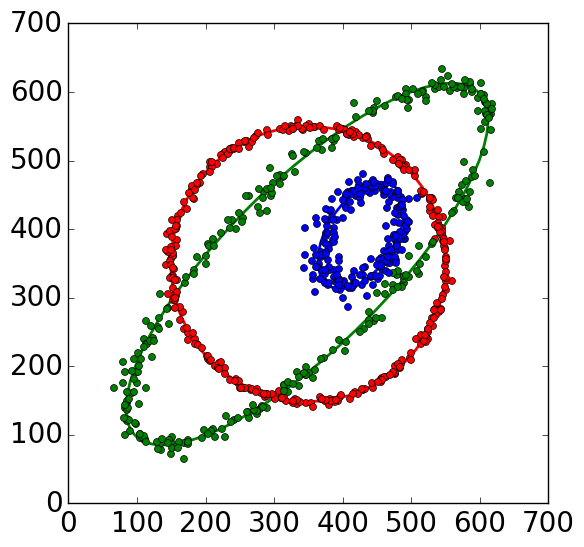}&
\includegraphics[scale=0.3]{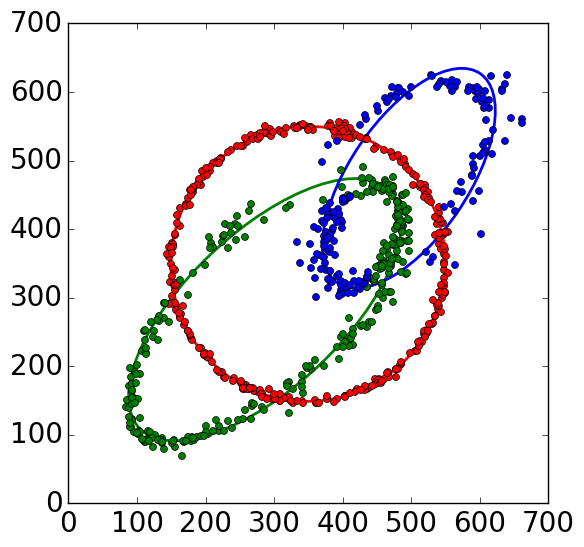}\\   
(e) & (f)
\end{tabular}
\caption{Synthetic 2D ellipse estimations. 
(a) Case 1: three ellipses and 350 outliers.
(b) Case 2: three ellipses and 800 outliers.
(c) Recovered four structures, case 1.
(d) Recovered first four structures, case 2.
(e) Three strongest structures, case 1.
(f) Interaction between two ellipses, case 2.}
\label{fig:synthellipse}
\end{figure}

\begin{figure}[t]
\centering
\begin{tabular}{@{\hspace{-0.0cm}}c@{\hspace{0.1cm}}c}
\includegraphics[scale=0.17]{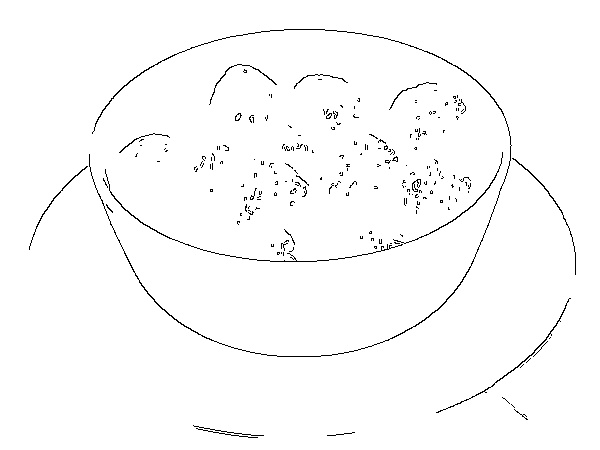}&
\includegraphics[scale=0.3]{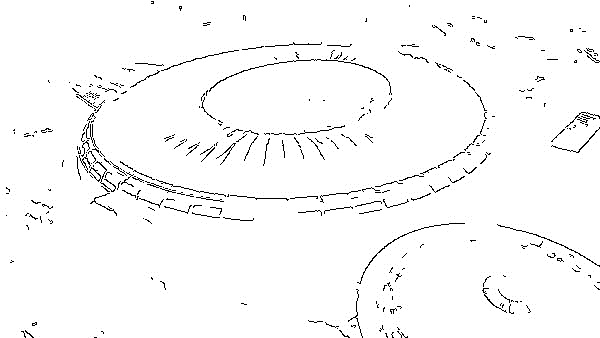}\\   
(a) & (b) \\
\includegraphics[scale=0.17]{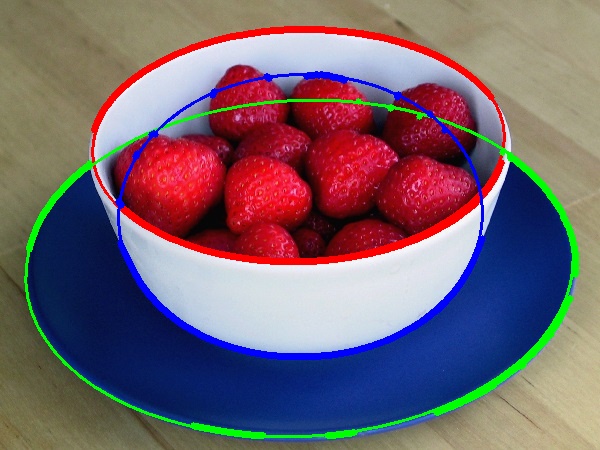}&
\includegraphics[scale=0.3]{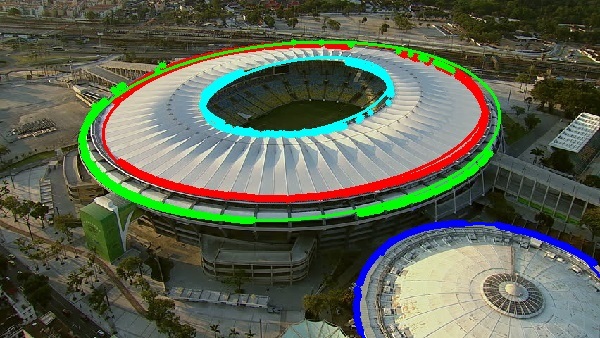}\\   
(c) & (d) 
\end{tabular}
\caption{2D ellipses in real images. 
(a) {\it Strawberries}: Canny edges, 4343 points.
(b) {\it Stadium}: Canny edges, 4579 points.
(c) {\it Strawberries}: Three strongest inlier structures.
(d) {\it Stadium}: Four strongest inlier structures (see also text).}
\label{fig:realellipse}
\end{figure}

In Fig.\ref{fig:synthellipse}a three ellipses are placed with 350 outliers
in the background. The inlier structures have
$n_{in} = 300, 250, 200$ and $\sigma_g = 3, 6, 9$.
The smallest ellipse with $n_{in} = 200$ is corrupted with the largest noise $\sigma_g = 9$.
We use $M = 5000$ in the ellipse fitting experiments.
When tested (Fig.\ref{fig:synthellipse}c), four ellipses are recovered 
\[\begin{array}{rcccc}
  & red  & green & blue & cyan\\
scale: & 12.1  & 28.9 & 48.0 & 1321.2\\
inliers: & 337 & 292 & 222 & 248\\
strength: & 28.0 & 10.1 & 4.6 & 0.2.
\end{array} \]
Based on results sorted by strength, 
the first three structures are inliers and
are shown in Fig.\ref{fig:synthellipse}e.

When the estimation is repeated 100 times, the three inlier 
structures are correctly located 97 times,
while in the other three tests 
the smallest ellipse is not estimated correctly.
From the 97 correct estimations, 
the average scales, the classified inlier amounts, 
along with their standard deviations are
\[\begin{array}{rccc}
scale: & 11.60 & 21.59 & 32.87 \\
& (1.54) & (3.64) & (14.71) \\
inliers: & 336.2 & 272.9 & 196.4 \\
& (8.2) & (27.5) & (53.2). \\
\end{array} \]
The average processing time is 3.28 seconds.

When the outlier amount reaches the limit,
the inlier structure with weakest strength may
no longer be sorted before the outliers.
The scale estimate becomes inaccurate due to the 
heavy outlier noise, and the outliers can form more dense 
structure with comparable strength.
When 800 outliers (Fig.\ref{fig:synthellipse}b) 
are placed in the image,
a test gives the result in Fig.\ref{fig:synthellipse}d.
The outlier structure (blue) has a strength of 4.8, while the value of 
the inliers (cyan) is 3.9.
However, the first two inlier structures are still recovered 
due to their stronger strengths. 

Fig.\ref{fig:synthellipse}b also gives an example to show 
one of the limitation explained in Section \ref{sec:inlieroutlier},
when the inlier strength is too weak to tolerate more outliers.
In Fig.\ref{fig:synthellipse}f two inlier structures interact, 
and the mean shifts converge to incorrect modes.

From Canny edge detection, 4343 and 4579 points are obtained 
from two real images containing several objects with elliptic shapes,
as shown in Fig.\ref{fig:realellipse}a and Fig.\ref{fig:realellipse}b.
With $M=5000$, the three strongest ellipses are drawn in 
Fig.\ref{fig:realellipse}c, superimposed over the original images.
The processing time is 18.90 seconds in this case.
In Fig.\ref{fig:realellipse}d the estimation takes 23.14 seconds to detect
four strongest ellipses, which are inlier structures. 
After 100 repetitive tests using the data shown 
in Fig.\ref{fig:realellipse}b, 
only the first two ellipses (red and green)
are detected reliably in 98 times.
The other two ellipses (blue and cyan) have smaller amounts of inliers
and therefore are less stable.
The data acquired from Canny edge detection
do not necessarily render the overall inlier structures in 
a more dense state. 
Preprocessing on the edge data is generally required for better performance.

\subsection{3D Cylinder}
\label{sec:3dcylinder}

The {\it 3D cylinders} estimation has the noisy objective function
\begin{equation}
\label{eqn:cylindereqn}
[\by_i ~ 1]\, \bP \, [\by_i ~ 1]^\top \simeq 0
\qquad i=1,\ldots,n_{in}
\end{equation}
where $\bP$ is a $4\times 4$ symmetric matrix.
The input variable $\by = [x ~ y ~ z]^\top$
gives the carrier vector
$\bx = [x^2 ~ xy ~ xz ~ y^2 ~ yz ~ z^2 ~ x ~ y ~ z]$. 
In the linear space there are nine degrees of freedom,
while a general cylinder is defined with only five degrees of freedom, four for the axis and one for the radius. 

A cylinder aligned with the z-axis has the equation
\begin{equation}
(x - s)^2 + (y - t)^2 - r^2 = 0 
\end{equation}
where $s,~t$ are the 2D coordinates of the axis
passing through the XY-plane, and $r$ is the radius.
These unknowns can be expressed in a quadric matrix
\begin{eqnarray}\nonumber
\bP' = \lambda \begin{bmatrix} \bD' & \bd' \\ \bd^{'T} & s^2 + t^2 -r^2 \end{bmatrix}
\qquad  \bD' = \begin{bmatrix} 1 & 0 & 0\\ 0 & 1 & 0 \\ 0 & 0 & 0 \end{bmatrix}\\
 \bd' = \begin{bmatrix} -s \\ -t \\ 0 \end{bmatrix} \qquad
 \bP = \begin{bmatrix} \bD & \bd \\ \bd^{T} & d \end{bmatrix} =
\bM^{-T} \bP' \bM^{-1}
\end{eqnarray}
with an euclidean transformation
 $\bM = \begin{bmatrix} \bR & \mathbf{t} \\ \mathbf{0}^{T} & 1 \end{bmatrix}$.
 The $3\times 3$ matrix $\bD$ remains singular, and has two identical eigenvalues.
 
Different approaches to compute the cylinder were given in \cite{Beder06}.
In this experiment, the nine-point method is used first to find 
a general quadric solution from each randomly initialized elemental subset, 
$\bA\, $vech$\bP = [$vech$^\top(X_i X_i^\top)]\ $vech$\bP = 0$,
i=1,...,9, where vech is the vectorization of the lower part of a symmetric matrix $\bP$, and $\bA$ a $9 \times 10$ matrix formed by stacking carrier vectors.
This approach gives an over-determined solution and $\bP$ could represent other quadrics instead of a cylinder.
The validity of a structure defined by an elemental subset 
can be checked with the singular values of
the $3\times 3$ matrix $\bD$, two of which should be quasi-equal
and the third close to zero, also $\bd$ is an eigenvector of $\bD$.

\begin{figure}[t]
\centering
\begin{tabular}{@{\hspace{-0.0cm}}c@{\hspace{-0.0cm}}c}
\includegraphics[scale=0.28]{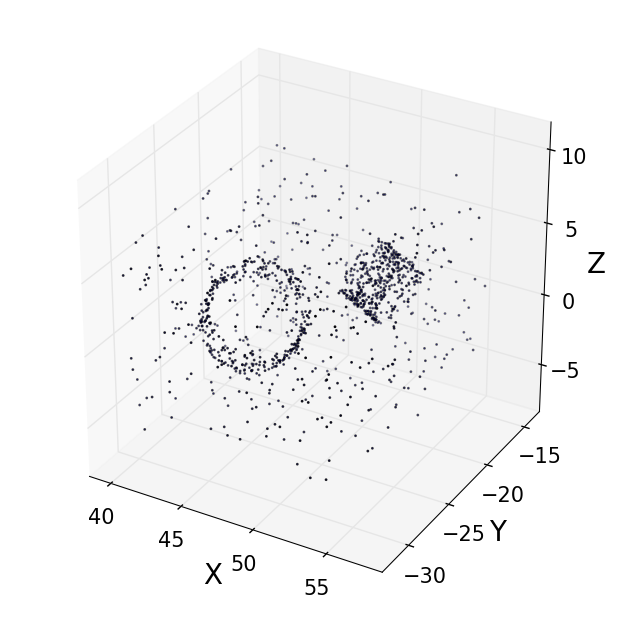}&
\includegraphics[scale=0.28]{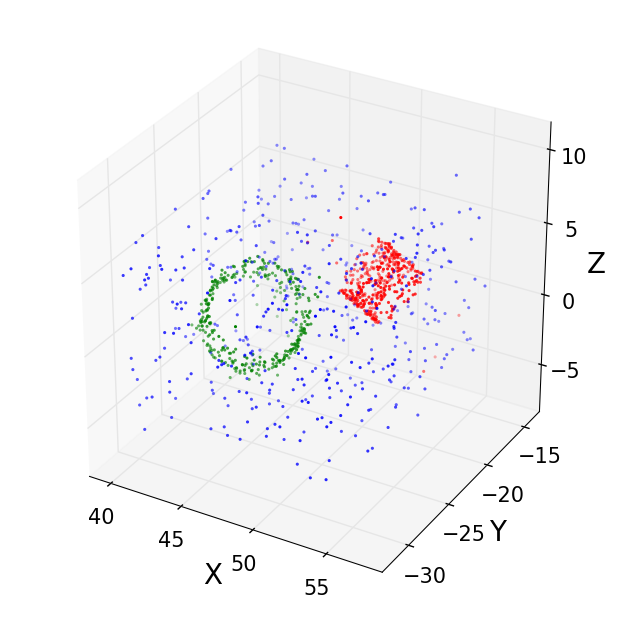} \\
(a) & (b)
\end{tabular}
\begin{tabular}{@{\hspace{-0.0cm}}c@{\hspace{-0.0cm}}c}
\includegraphics[scale=0.28]{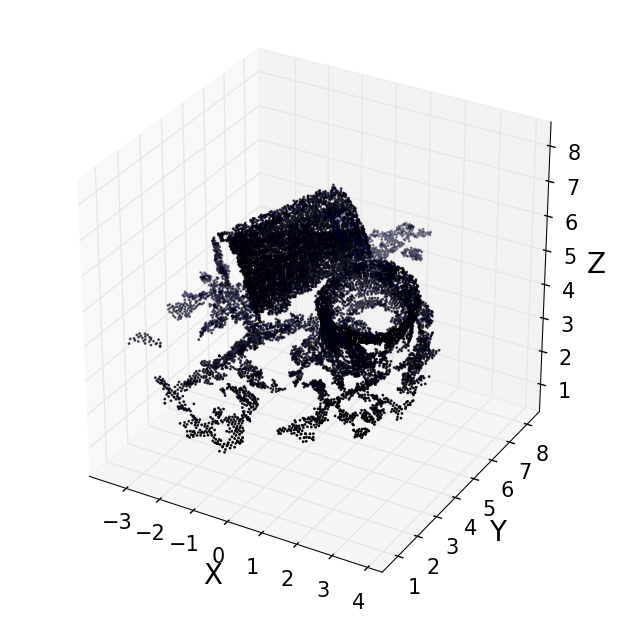}&
\includegraphics[scale=0.28]{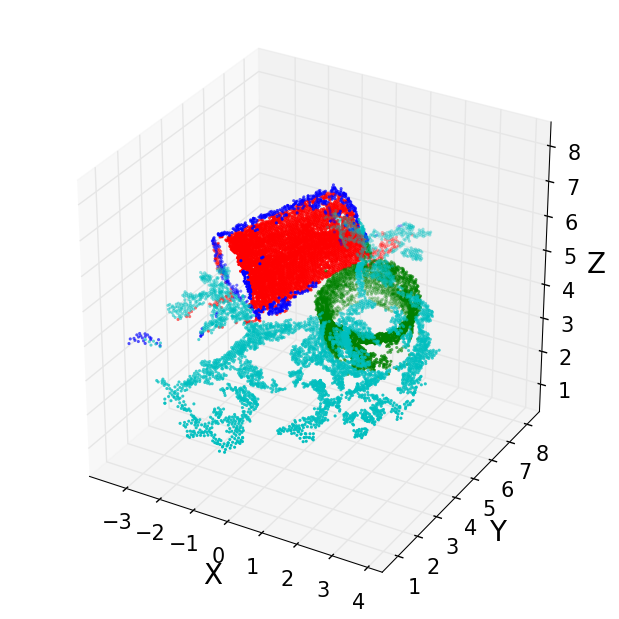}\\   
(c) & (d) 
\end{tabular}
\caption{3D cylinder estimations. 
(a) Two synthetic cylinders and 500 outliers.
(b) Two inlier and one outlier structures are recovered.
(c) The 3D cloud of input points.
(d) Three inlier and one outlier structures are recovered.}
\label{fig:cylinder}
\end{figure}

The transpose of the $9\times 3$ Jacobian matrix is
\begin{equation}
\label{eqn:cylinderjacob}
\bJ_{\scriptsize{\mathbf{x} | \mathbf{y}}}^\top = 
\left[ \begin{array}{ccccccccc} 
2x & y & z & 0 & 0 & 0 & 1 & 0 & 0\\
0 & x & 0 & 2y & z & 0 & 0 & 1 & 0\\
0 & 0 & x & 0 & y & 2z & 0 & 0 & 1
\end{array} \right] .
\end{equation}
In Fig.\ref{fig:cylinder}a two cylinders 
are placed with 500 outliers.
Each cylinder has its rotation axis in a randomly generated direction.
The inlier structures have radius $r = 2, 3$, the number of inliers
$n_{in} = 400, 300$ and $\sigma_g = 0.06, 0.1$.
The i.i.d. inlier noise is applied in 3D dimension, which is 
roughly $\pm10\%$ of the size of radius.

With $M = 5000$ the result becomes stable and three strongest structures are returned
\[\begin{array}{rccc}
  & red  & green & blue\\
scale: & 0.28  & 0.48 & 5.56\\
inliers: & 413 & 337 & 449\\
strength: & 1487.1 & 705.0 & 80.8.
\end{array} \]
in 25.02 seconds.
The first two structures are inliers and 
the first outlier structure has a much weaker strength, 
as shown in Fig.\ref{fig:cylinder}b.
After testing the randomly generated data for 100 trials,
in 96 trials the stronger cylinder is correctly segmented,
while the weaker one in 94 trials.

The {\it VisualSFM} software \cite{Wu13, Furu10} 
is used to generate the point cloud of 
a 3D scene captured by an image sequence.
The point cloud consists of 12103 points (Fig.\ref{fig:cylinder}c). 
Compared with the synthetic data,
the inliers in the point cloud are more dense and 
have much smaller noise.
A smaller sampling size $M = 2000$ gives a stable result.
The processing takes 44.75 seconds
and as Fig.\ref{fig:cylinder}d shows,
the three inlier structures 
(red, green and blue) and the first outlier structure (cyan)
are recovered.
The heights of the cylinders are not evaluated in
these tests, which would require post-processing. 

\subsection{Fundamental matrix}
\label{sec:fundmat}

The next experiment shows the estimation of the {\it fundamental matrices}. 
The corresponding linear space was introduced in Section \ref{sec:transformation}. 
The $3\times 3$ matrix $\bF$ is rank-2 and 
a recent paper \cite{cheng15} solved the 
non-convex problem iteratively 
by a convex moments based polynomial optimization.
It compared the results with a few RANSAC type algorithms.
The method required several parameters and
in each example only one fundamental matrix was recovered.

\begin{figure}[t]
\centering
\begin{tabular}{@{\hspace{-0.0cm}}c@{\hspace{-0.0cm}}c}
\includegraphics[scale=0.3125]{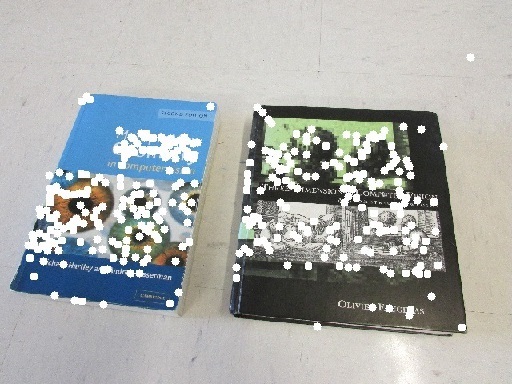} &
\includegraphics[scale=0.3125]{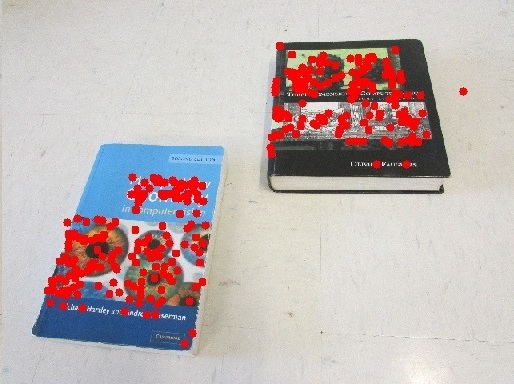}\\
(a) & (b) \\ 
\includegraphics[scale=0.3125]{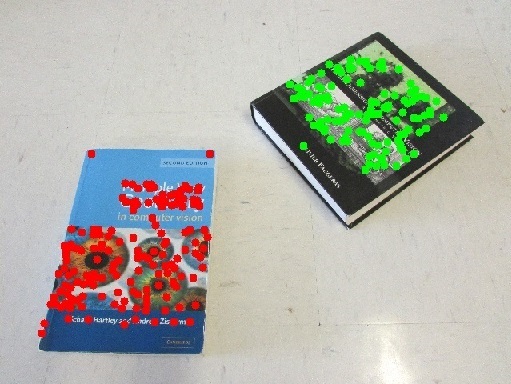}&
\includegraphics[scale=0.25]{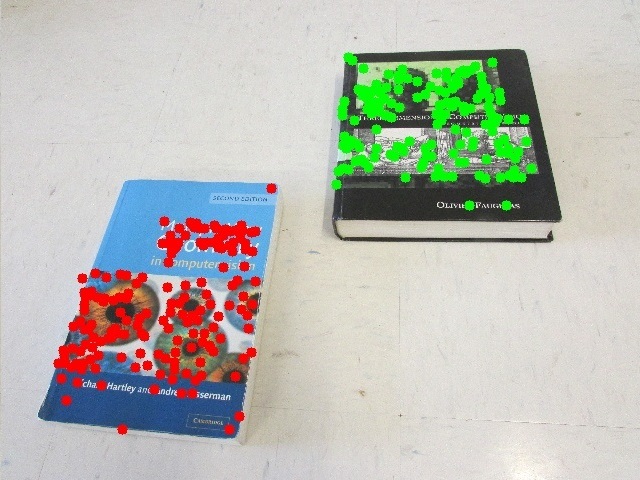}\\
(c) & (d)
\end{tabular}
\caption{Motion segmentation with different objective functions. 
In the following figures, the input points (white) are shown in the
first view, the processed structures (colored) in the second view.
(a) The input points.
Fundamental matrix: (b) Translation only. (c) Translation and rotation.
Homography: (d) Translation only.}
\label{fig:fundmatFig1}
\end{figure}

The fundamental matrix cannot be used to segment objects 
with only translational motions, as proved in \cite{babHadiashar09}.
The input shown in 
Fig.\ref{fig:fundmatFig1}a comes out in Fig.\ref{fig:fundmatFig1}b as a single structure instead of two. 
Only when one of the books has a large enough rotation, the correct output is obtained (Fig.\ref{fig:fundmatFig1}c).
Applying the homography estimation (Section \ref{sec:2dhomography}) to the translational case,  
the two books can be easily separated (Fig.\ref{fig:fundmatFig1}d). 
However, this problem was not explicitly mentioned when quasi-translational images were used for testing as in \cite{pham14}.

\begin{figure}[h]
\centering
\begin{tabular}{@{\hspace{-0.0cm}}c@{\hspace{-0.0cm}}c}
\includegraphics[scale=0.3125]{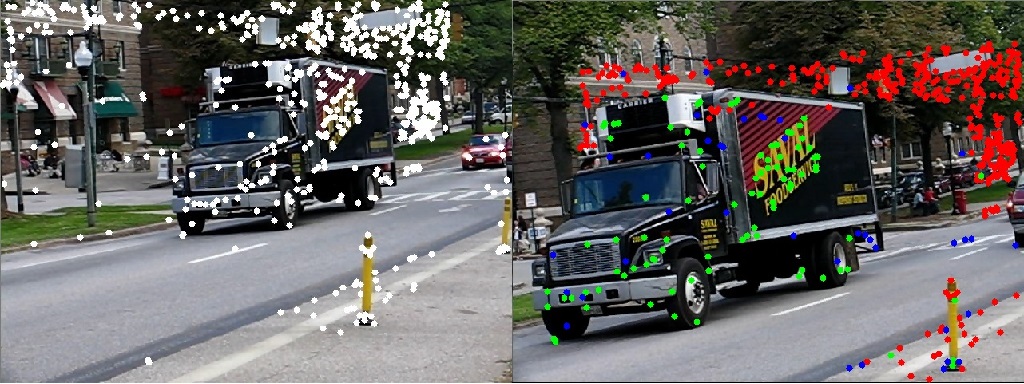}\\
(a) \\ 
\includegraphics[scale=0.3125]{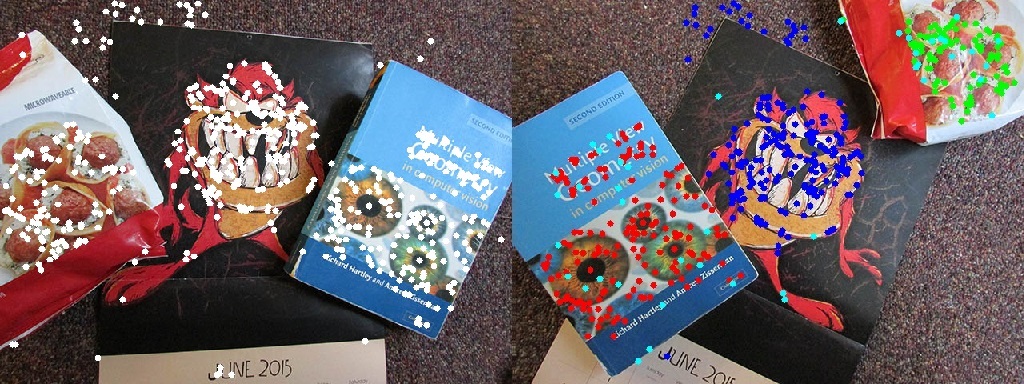}\\   
(b) \\
\includegraphics[scale=0.3125]{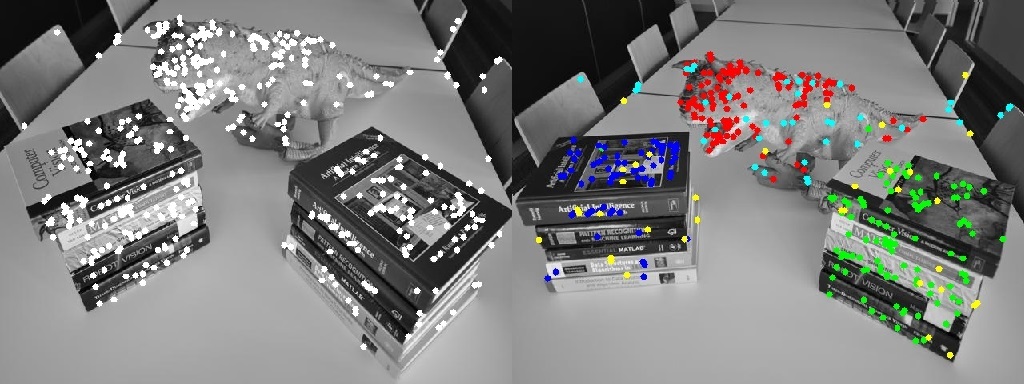}\\   
(c)
\end{tabular}
\caption{Fundamental matrix estimation. 
(a) Image pair from {\it Hopkins 155 dataset} with two inlier
and one outlier structures.
(b) The {\it books} with three inlier and one outlier structures.
(c) The {\it dinabooks} from \cite{pham14} with 
four inlier and one outlier structures.}
\label{fig:fundmatFig2}
\end{figure}

Each example of Fig.\ref{fig:fundmatFig2} shows the movement of multiple rigid objects.
The point correspondences are extracted by
OpenCV with a distance ratio of 0.8 for SIFT \cite{lowe04},
giving 608, 614 and 457 matches, respectively.
With $M=5000$, the structures are retained as
\[\begin{array}{rccc}
\mbox{Fig.\ref{fig:fundmatFig2}a}  & red  & green & blue\\
scale: & 0.56  & 0.73 & 11.78\\
inliers: & 407 & 101 & 51\\
strength: & 727.3 & 139.3 & 4.33.
\end{array} \]
\[\begin{array}{rcccc}
\mbox{Fig.\ref{fig:fundmatFig2}b}  & red  & green & blue & cyan\\
scale: & 0.46  & 0.40 & 1.12 & 10.42\\
inliers: & 192 & 96 & 221 & 47\\
strength: & 413.4 & 242.8 & 196.8 & 4.5.
\end{array} \]
\[\begin{array}{rccccc}
\mbox{Fig.\ref{fig:fundmatFig2}c}  & red  & green & blue & cyan & yellow\\
scale: & 0.22  & 0.72 & 0.65 & 0.70 & 23.9\\
inliers: & 135 & 117 & 84 & 48 & 43\\
strength: & 623.0 & 161.5 & 129.0 & 68.2 & 1.8.
\end{array} \]
The estimations take 1.75, 2.30 and 2.10 seconds for these three cases.
In real images, the outlier structures can be easily filtered out since they
have much larger scales than the inliers. 
It can be observed that the scales of the inlier structures are very close,
therefore the methods with fixed thresholds may be used here. 
However, if the images are scaled before estimation, the error in the inliers will change proportionally. Correct scale estimate can only be found adaptively from the input data.

As discussed in Section \ref{sec:inlieroutlier},
the first (red) and the fourth (cyan) structures obtained
from Fig.\ref{fig:fundmatFig2}c can be fused as a single structure.
This merge has to be done by post-processing in the input space but
also requires a threshold from the user.

The SIFT matches are error-prone and
false correspondences always exist.
If the images contain repetitive features, 
such as the exterior of buildings,
parametrization of the repetitions can reduce the uncertainty
\cite{Schaffalitzky99}. 
Preprocessing of the images is not described 
in this paper, therefore we will not explain it further.

\subsection{Homography}
\label{sec:2dhomography}

\begin{figure}[t]
\centering
\begin{tabular}{c}
\includegraphics[scale=0.25]{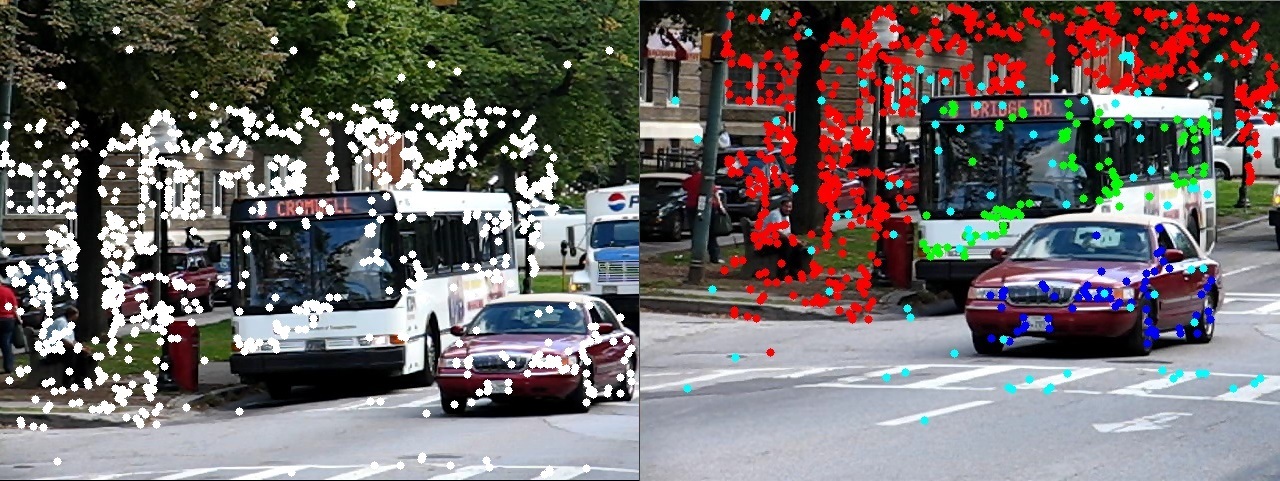}\\
(a) \\
\includegraphics[scale=0.5]{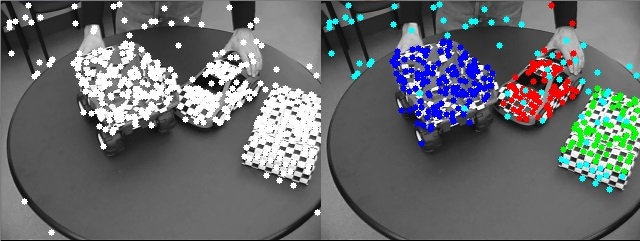}\\
(b)
\end{tabular}
\caption{Homography estimation with
image pairs from {\it Hopkins 155} dataset. 
(a) Three inlier and one outlier structures.
(b) Three inlier and one outlier structures.}
\label{fig:homographyFig1}
\end{figure}

\begin{figure}[t]
\centering
\begin{tabular}{@{\hspace{-0.0cm}}c@{\hspace{-0.0cm}}c}
\includegraphics[scale=0.3125]{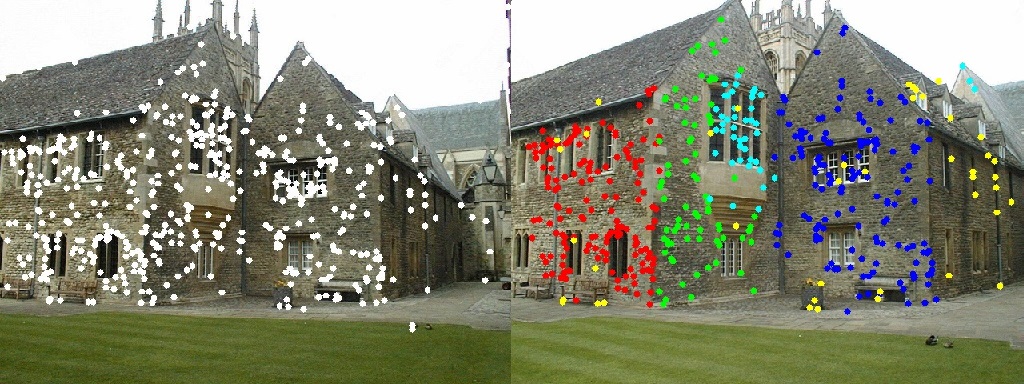}\\   
(a) \\
\includegraphics[scale=0.3125]{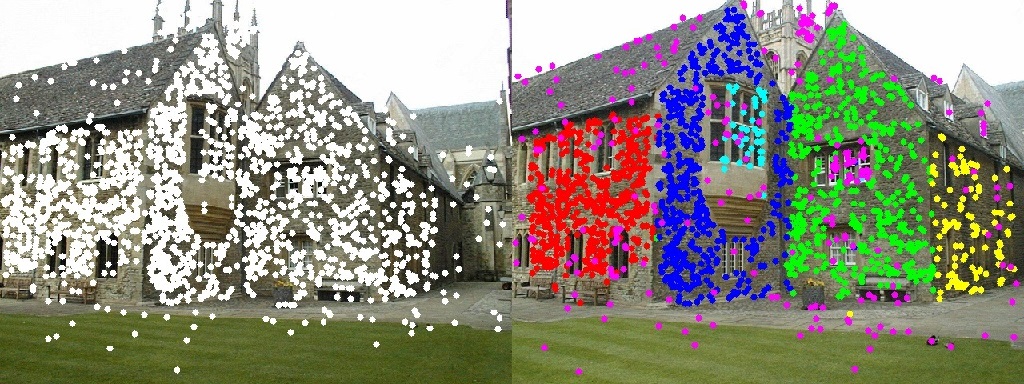}\\
(b) \\ 
\includegraphics[scale=0.3125]{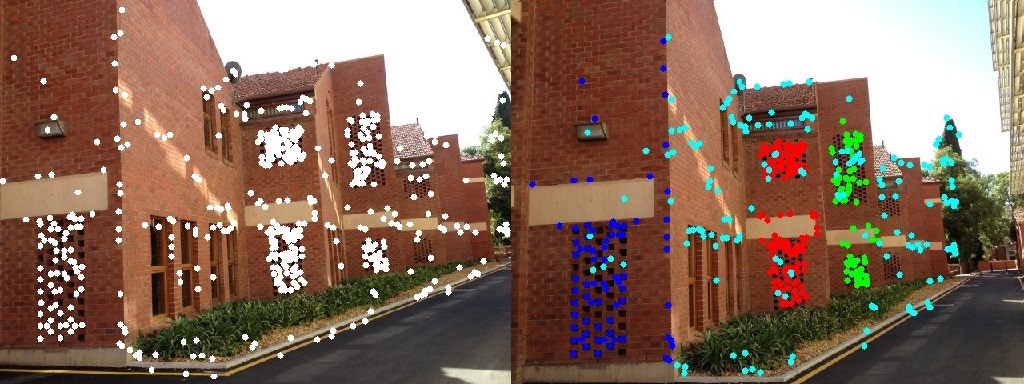}\\   
(c) \\
\includegraphics[scale=0.3125]{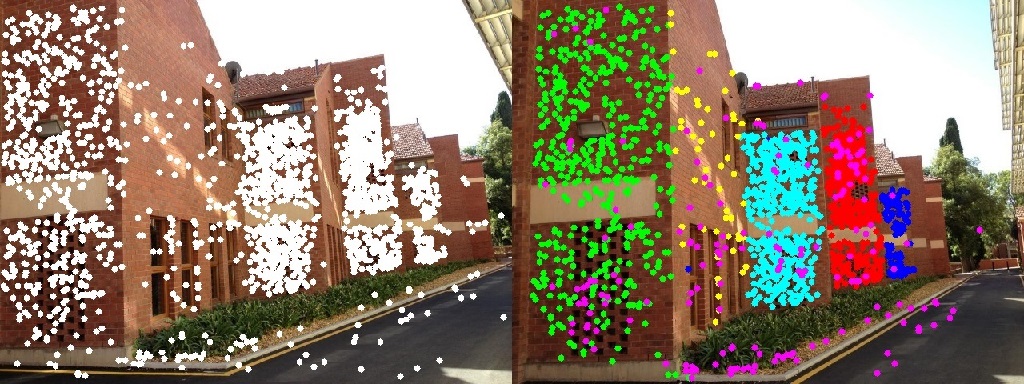}\\   
(d)
\end{tabular}
\caption{Homography estimation.
(a) {\it Merton College} with 536 point pairs. Four inlier and one outlier structures.
(b) {\it Merton College} with 1982 point pairs. Five inlier and one outlier structures.
(c) {\it Unionhouse} with 619 point pairs. Three inliers and one outlier structures.
(d) {\it Unionhouse} with 2084 point pairs. Five inliers and one outlier structures.}
\label{fig:homographyFig2}
\end{figure}

The final example is for 2D {\it homography} estimation.
Each inlier structure is represented by a $3\times 3$ matrix $\bH$, 
which connects two planes inside the image pair
\begin{equation}
\label{eqn:homography}
\mathbf{y}_{i}^\prime \simeq \bH \by_i,\quad i = 1,\ldots,n_{in}
\end{equation}
where $\by = [x ~ y ~ 1]^\top$ and 
$\by^\prime = [x^\prime ~ y^\prime ~ 1]^\top$ are the homogeneous
coordinates in these two images. 

As mentioned in Section \ref{sec:prepare}, 
the homography estimation has $\zeta =2$. 
The input variables are $[x~~y~~x^\prime~~y^\prime]^\top$.
Two linearized relations can be derived from the
constraint (\ref{eqn:homography}) by the direct linear transformation (DLT)
\begin{equation}
\label{eqn:DLThomo}
\mathbf{A}_i\mathbf{h} = \left[
\begin{array}{@{\hspace{-0.03cm}}c@{\hspace{0.1cm}}c@{\hspace{0.1cm}}c@
{\hspace{-0.00cm}}}
-\mathbf{y}_i^{\top} & \phantom{-}\mathbf{0}^{\top}_3 &
\phantom{-}x_{i}^{\prime}\mathbf{y}_i^{\top} \\
\mathbf{0}^{\top}_3  & -\mathbf{y}_i^{\top} &
\phantom{-}y_{i}^{\prime}\mathbf{y}_i^{\top} \\
\end{array}\right]
\left[\begin{array}{@{\hspace{-0.03cm}}c@{\hspace{-0.03cm}}}
\mathbf{h}_1\\ \mathbf{h}_2\\ \mathbf{h}_3\\
\end{array} \right]
\simeq \mathbf{0}_2~.
\end{equation} 
The matrix $\bA_i$ is $2\times 9$ and both rows satisfy 
the relations with the vector derived from the matrix 
vec$(\bH^\top) = \bh = \btheta$.

The carriers are obtained from the two rows of $\bA_i$
\begin{eqnarray}
\label{eqn:carriershomography}
\bx^{[1]} &=& [ {-x} ~~ {-y} ~~ {-1} ~~ 0 ~~ 0 ~~ 0 ~~ x^\prime x ~~ 
x^\prime y ~~ x^\prime ]^\top \nonumber \\
\bx^{[2]} &=& [ 0 ~~ 0 ~~ 0 ~~  {-x} ~~ {-y} ~~ {-1} ~~ y^\prime x ~~ 
y^\prime y ~~ y^\prime ]^\top .
\end{eqnarray}
The transpose of the two $9\times 4$ Jacobians matrices are
\begin{align}
\nonumber\mathbf{J}_{\scriptsize{\mathbf{x}_i^{[1]}|\mathbf{y}}}^\top &=
\left[
\begin{tabular}{@{\hspace{-0.01cm}}c@{\hspace{-0.01cm}}c@
{\hspace{-0.001cm}}c@{\hspace{-0.01cm}}} {$-\mathbf{I}_{2\times 2}$} &
\multirow{3}{*}{$\phantom{-}\mathbf{0}_{4\times 4}$} &
$\phantom{-}x^\prime_i\mathbf{I}_{2\times 2}~~\mathbf{0}_2$\\
{$\phantom{0}\mathbf{0}_{2}^\top$} & & $\phantom{--}\mathbf{y}_i^\top$\\
{$\phantom{0}\mathbf{0}_{2}^\top$} & &
{$\phantom{--}\mathbf{0}_{2}^\top~~~~0$}\\
\end{tabular} \right]\\
\mathbf{J}_{\scriptsize{\mathbf{x}_i^{[2]}|\mathbf{y}}}^\top &=
\left[\begin{tabular}{@{\hspace{-0.01cm}}ccc@{\hspace{-0.01cm}}c@
{\hspace{-0.01cm}}} \multirow{3}{*}{$\mathbf{0}_{4\times 3}$} &
$-\mathbf{I}_{2\times 2}$ & \multirow{3}{*}{$\mathbf{0}_{4}$} &
{$\phantom{0}y^\prime_i\mathbf{I}_{2\times 2}~~~\mathbf{0}_2$} \\
& {$\mathbf{0}_{2}^\top$} & & $\phantom{-}\mathbf{0}_{2}^\top~~~~~0$ \\
& {$\mathbf{0}_{2}^\top$} & & $\phantom{--}\mathbf{y}_i^\top$ \\
\end{tabular}
\right] .
\end{align}
Based on Section \ref{sec:reducedistance}, for every $\btheta$
only the larger Mahalanobis distance is used for each
input $\by_i, \; i=1,\ldots,n$.

The motion segmentation involves only translation in 3D in
Fig.\ref{fig:homographyFig1}a and Fig.\ref{fig:homographyFig1}b,
both images are taken from the {\it Hopkins 155} dataset.
With $M=2000$, 
the processing time is 1.12 and 1.09 seconds for the inputs containing 990 and 482 SIFT point pairs, respectively.
As mentioned in Section \ref{sec:inlieroutlier}, in 
Fig.\ref{fig:homographyFig1}a 
the estimator cannot separate these two 3D planes on the bus
because the 2D homographies corresponding to them are very similar.
The condition (\ref{eqn:condition}) does not stop where it should 
since the points on either 2D planes are not dense enough.
The example shows that the desired results can only be obtained by increasing the amount of inliers from preprocessing.
In Fig.\ref{fig:homographyFig1}b, the three objects can
be correctly separated in spite of the very small motions, 
due to the stronger strengths in all the inlier structures.

\[\begin{array}{rcccc}
\mbox{Fig.\ref{fig:homographyFig1}a}  & red  & green & blue & cyan\\
scale: & 1.62  & 1.12 & 4.49 & 203.11\\
inliers: & 713 & 101 & 67 & 105\\
strength: & 440.2 & 89.5 & 14.9 & 0.5
\end{array} \]
\[\begin{array}{rcccc}
\mbox{Fig.\ref{fig:homographyFig1}b}  & red  & green & blue & cyan\\
scale: & 0.20  & 0.17 & 0.56 & 5.16\\
inliers: & 107 & 88 & 165 & 89\\
strength: & 529.6 & 517.7 & 293.4 & 17.3.
\end{array} \]

The importance of preprocessing can also be seen in \cite{serradell10},
where the geometric and appearance priors were used to increase the 
amount of consistent matches before estimation,
when the PROSAC \cite{chum05} failed in the presence of many incorrect matches.

The significance of inlier amounts is demonstrated
in Fig.\ref{fig:homographyFig2}.
The point pairs from OpenCV SIFT are used in 
Fig.\ref{fig:homographyFig2}a and Fig.\ref{fig:homographyFig2}c,
while the more dense points from datasets \cite{pham14} are tested in 
Fig.\ref{fig:homographyFig2}b and Fig.\ref{fig:homographyFig2}d for 
comparison.
After the estimations, the structures are sorted by their
strengths until the first outlier structure appears, 
where a large increase in the scale estimate can be observed. 

\[\begin{array}{rccccc}
\mbox{Fig.\ref{fig:homographyFig2}a}  & red  & green & blue & cyan & yellow\\
scale: & 1.41  & 0.89 & 2.01 & 1.37& 14.26\\
inliers: & 157 & 87 & 158 & 51 & 44\\
strength: & 110.8 & 97.2 & 78.5 & 37.0 & 3.1
\end{array} \]
\[\begin{array}{rcccccc}
\mbox{Fig.\ref{fig:homographyFig2}b}  & red  & green & blue & cyan & yellow & purple\\
scale: & 0.97  & 0.92 & 2.55 & 0.41& 0.97 & 798.0\\
inliers: & 507 & 478 & 529 & 55 & 127 & 286\\
strength: & 521.5 & 518.9 & 207.6 & 133.3 & 130.1 & 0.4
\end{array} \]
\[\begin{array}{rcccc}
\mbox{Fig.\ref{fig:homographyFig2}c}  & red  & green & blue & cyan\\
scale: & 0.59  & 0.48 & 0.74 & 60.34\\
inliers: & 176 & 87 & 91 & 204\\
strength: & 295.9 & 181.0 & 123.3 & 3.4
\end{array} \]
\[\begin{array}{rcccccc}
\mbox{Fig.\ref{fig:homographyFig2}d}  & red  & green & blue & cyan & yellow & purple\\
scale: & 0.67  & 1.35 & 0.57 & 1.83 & 1.53 & 82.18\\
inliers: & 495 & 508 & 162 & 513 & 69 & 210\\
strength: & 738.7 & 376.7 & 282.3 & 280.1 & 44.9 & 2.6.
\end{array} \]

The processing time for these four examples is 1.29, 3.07, 1.36 and 3.78
seconds, respectively.
This is a faster implementation than RCMSA \cite{pham14}
which cost 24.58 and 25.40 sec for the same image pairs in Fig.\ref{fig:homographyFig2}b and Fig.\ref{fig:homographyFig2}d on the same computer.
With  denser inlier points, more inlier structures can be detected.

\section{Discussion}
\label{sec:discussion}

A new robust algorithm was presented which does not require the 
inlier scales to be  specified by the user prior the estimation.
It estimates the scale for each structure adaptively from the input data,
and can handle inlier structures with different noise levels.
Using a  strength based classification, 
the inlier structures with larger strengths are retained 
while a large quantity of outliers are removed. 

In Section \ref{sec:robustness} we will summarize the
conditions for robustness, 
and in Section \ref{sec:unsolved} several 
open problems will be reviewed.

\subsection{Conditions for robustness}
\label{sec:robustness}

We have discussed several limitations for the 
algorithm in Section \ref{sec:inlieroutlier}
without considering the number of trials $M$
for random sampling, the 
only parameter given by the user.

In the new algorithm $M$ trials are used to estimate the scale.
The required amount of $M$ depends strongly on the data to be processed.
The complexity of the objective function,
the size of the input data, the number of inlier structures, 
the inlier noise levels, and the amount of outliers,
all are factors which can affect the required number of trials.
If no information on the size of $M$ is known, 
the user can run several tests with different $M$-s until the results become stable. Once $M$ is large enough, the estimation will not improve by using larger sampling size.

With all the other robustness conditions satisfied, $M$
has to be large enough to detect the weakest inlier structure.
This process gives a quasi-correct scale estimate 
and then the mean shift recovers a desired inlier structure.
Once the interaction between inliers and outliers is apparent,
the quality of the estimation cannot really be compensated by a larger $M$ since the initial set has become less reliable. 
Only through preprocessing of the input data will the number of inlier points increase and a better result be obtained.

Three main conditions to improve 
the robustness of the proposed algorithm are summarized:
\begin{itemize}
\item Preprocessing to reduce the amount of outliers, 
while bring in more inliers.
\item The sampling size $M$ should be large enough to stably find the inlier estimates.
\item Post-processing should be done in the input space
when an inlier structure comes out split or has to be separated.
\end{itemize} 

\subsection{Open problems}
\label{sec:unsolved}

We will list several open problems and
discuss possible solutions, where further research and experiments 
are still needed.

Assume that the measurements of the input points
have different covariances which are not specified.
In many computer vision problems this situation 
is neglected but can still exist. 
The homoscedastic inlier covariances have the form
\begin{equation}
\label{eqn:covmulti}
[ \sigma_1^2 \ldots \sigma_{\scriptsize{\by}}^2] \bI_{\scriptsize{\by}} =
 \begin{bmatrix} \sigma_1^2 & \cdots & 0 \\
\cdots & \cdots & \cdots \\
0 & \cdots & \sigma_{\scriptsize{\by}}^2 \end{bmatrix}
\end{equation}
with $\sigma_j$-s unknown. 
The computation for the covariance of the carrier
(\ref{eqn:newcovariance}) places the $\sigma_j$-s
into the product of two Jacobian matrices.

A possible solution is to start with a uniform $\sigma$, 
that is, $\sigma^2\bI_{\by}$. 
Normally the algorithm should work,
but each inlier structure may not attract the 
quasi-correct amount of points. 
Then for each inlier structure separately,
consider a local region where the inliers are located.
This region can be relatively larger to include additional inliers and outliers.
Apply the entire estimation again only on the data inside this region,
in this way a more accurate $\hat{\sigma}_j$ could be found and 
to update the final estimate.

In  face image classification or projective motion factorization,
the objective functions have only one carrier vector $\zeta = 1$, but the
estimate is a $m\times k$ matrix $\bTheta$ and 
a $k$-dimensional vector $\balpha$. 
Since $\zeta$ is one dimensional, here we
use $\bx$ instead of $\tilde\bx$.
The covariance of $\bz_i$ is
$\sigma^2\bH_i = \sigma^2\bTheta^\top \bC_i \bTheta$, 
with $\sigma^2$ unknown.
This gives a $k\times k$ symmetric Mahalanobis distance matrix 
for $i=1,\ldots,n$
\begin{equation}
\label{eqn:matrix}
\bD_i = \sqrt{ \left( \bx_i^\top \bTheta - 
\balpha \right)^\top \bH_i^{-1} 
\left( \bx_i^\top \bTheta - \balpha \right)}
\end{equation}
which could be expressed as the union of $k$ vectors 
\begin{equation}
\label{eqn:svdmahalanobis}
\bD_i = [\bd_{i:1}~\ldots~ \bd_{i:k}] .
\end{equation}
A possible solution is to 
order the Mahalanobis distances $\bd_{[i:*]}$
for each column separately,
and collect the inputs corresponding to the minimum sum of 
distances for $\epsilon\%$ of the data.
The $k\times k$ matrices are reduced to $k$ initial sets, 
one for each dimension. 

Apply independently $k$ times the
expansion process described in Section \ref{sec:newsigma} and define the 
$k\times k$ diagonal scale matrix
\begin{equation}
\label{eqn:multiplesigma}
\bS_k = \begin{bmatrix} \hat{\sigma}_1 & \cdots & 0 \\
\cdots & \cdots & \cdots \\
0 & \cdots & \hat{\sigma}_k \end{bmatrix} .
\end{equation}
The $k\times k$ covariance matrix $\bB_i$ is computed as
\begin{equation}
\label{eqn:multiband}
\bB_i = \bS_k^\top \, \bTheta^\top \bC_i \bTheta \, \bS_k .
\end{equation}
The second step in the algorithm, the mean shift, is now multidimensional and further experiments will be needed to verify 
the feasibility of this solution.

If an image contains both lines and conics, 
there is no clear way to estimate both of them properly.
See examples Fig.\ref{fig:realline}c and Fig.\ref{fig:realline}d.
Supposing we start with the lines, 
then some points from the conics 
may also be classified as lines. 
However, these points should be 
put back into the input data and used for the conics,
otherwise some conics may not be detected. 
The same problem arises when we estimate the conics first.
Only through supplemental processing we may possibly 
separate the line from the conic structures.

The Python/C++ program for the robust estimation of 
multiple inlier structures is posted on our website at \\
{\tt \centerline{coewww.rutgers.edu/riul/research/code/}
\centerline{MULINL/index.html.}}

\bibliographystyle{alphaieeetr}
\bibliography{ourreference}

\vspace{1cm}
\begin{wrapfigure}{l}{0.15\textwidth}
\includegraphics[width= 1.1in]{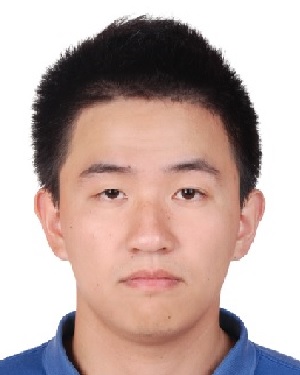}
\end{wrapfigure}
\textbf{Xiang Yang} received the BE degree in
mechanical engineering and automation
from Beihang University, Beijing, China, in 2009, 
and the MS degree in mechanical engineering 
from University of Bridgeport, Connecticut, in 2012. 
Currently, he is working toward the PhD degree in mechanical and
aerospace engineering at Rutgers University, New Jersey. 
His research interests include computer aided design,
3D reconstruction and statistical pattern recognition.

\vspace{0.5cm}
\begin{wrapfigure}{l}{0.15\textwidth}
\includegraphics[width= 1.1in]{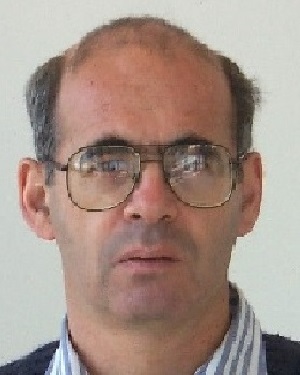}
\end{wrapfigure}
\textbf{Peter Meer} received the Dipl. Engn. degree from the Bucharest
Polytechnic Institute, Romania, in 1971, and the D.Sc.
degree from the Technion, Israel Institute of Technology, Haifa,
in 1986, both in electrical engineering.
From 1971 to 1979 he was with the Computer Research Institute,
Cluj, Romania, and
between 1986 and 1990 with
Center for Automation Research, University of Maryland at
College Park. In 1991 he joined  the Department of Electrical
and Computer Engineering, Rutgers University, NJ
and is currently a Professor.
He has held visiting appointments in Japan, Korea, Sweden, Israel
and France.
He was an Associate Editor of the {\it IEEE Transaction on Pattern
Analysis and Machine Intelligence} between 1998 and 2002.
With coautors Dorin Comaniciu and Visvanathan Ramesh he
received at the 2010 CVPR the Longuet-Higgins prize for fundamental
contributions in computer vision.
His research interest is in application of modern statistical methods
to image understanding problems. He is an IEEE Fellow.

\end{document}